\documentclass[pdflatex,sn-mathphys-num]{sn-jnl}% Math and Physical Sciences Numbered Reference Style
\usepackage{amssymb}
\usepackage{amsmath}
\usepackage{algorithmic}
\usepackage{graphicx}
\usepackage{textcomp}
\usepackage{multirow}
\usepackage{longtable}
\usepackage{array}
\usepackage{url}
\usepackage{tabularx}
\usepackage{colortbl}
\usepackage{booktabs}  
\usepackage{acronym}
\usepackage{xcolor}

\acrodef{AI}{Artificial Intelligence}
\acrodef{ANN}{Artificial Neural Network}
\acrodef{ARM}{Advanced RISC Machine}
\acrodef{AutoML}{Automated Machine Learning}
\acrodef{BCNN}{Binary Convolutional Neural Network}
\acrodef{BERT}{Bidirectional Encoder Representations from Transformers}
\acrodef{BLE}{Bluetooth Low Energy}
\acrodef{BMLP}{Binary Multi-Layer Perceptron}
\acrodef{BNN}{Binary Neural Network}
\acrodef{CLA}{Custom Latch Array}
\acrodef{CLADIMC}[CLA-DIMC]{Custom Latch Array Digital In-Memory Computing}
\acrodef{CMSISNN}[CMSIS-NN]{Cortex Microcontroller Software Interface Standard Neural Network}
\acrodef{CNN}{Convolutional Neural Network}
\acrodef{CPU}{Central Processing Unit}
\acrodef{DIMC}{Digital In-Memory Computing}
\acrodef{DSCNN}{Depthwise Separable Convolutional Neural Network}
\acrodef{DT}{Decision Tree}
\acrodef{EdgeGenAI}[Edge-GenAI]{Edge Generative Artificial Intelligence}
\acrodef{EdgeML}{Edge Machine Learning}
\acrodef{EIT}{Electrical Impedance Tomography}
\acrodef{EKF}{Extended Kalman Filter}
\acrodef{FANN}{Fast Artificial Neural Network}
\acrodef{FC}{Fully Connected}
\acrodef{FCN}{Fully Connected Network}
\acrodef{FIR}{Finite Impulse Response}
\acrodef{FL}{Federated Learning}
\acrodef{FPGA}{Field-Programmable Gate Array}
\acrodef{HAL}{Hardware Abstraction Layer}
\acrodef{HAR}{Human Activity Recognition}
\acrodef{HBA}{Human Behavior Analysis}
\acrodef{IoMT}{Internet of Medical Things}
\acrodef{IoT}{Internet of Things}
\acrodef{IUAV}{Intelligent Unmanned Aerial Vehicle}
\acrodef{KNN}{K-Nearest Neighbors}
\acrodef{KNNEM}[KNN-EM]{K-Nearest Neighbors Expectation-Maximization}
\acrodef{LCROpt}[LCR-Opt]{Optimized Locality-Constrained Regularized}
\acrodef{LSTM}{Long Short-Term Memory}
\acrodef{MAE}{Mean Absolute Error}
\acrodef{MASLR}[MA-SLR]{Mean Adjusted Stochastic Linear Regression}
\acrodef{MCU}{Microcontroller Unit}
\acrodef{MFCC}{Mel-Frequency Cepstral Coefficient}
\acrodef{ML}{Machine Learning}
\acrodef{MLP}{Multi-Layer Perceptron}
\acrodef{NB}{Naive Bayes}
\acrodef{NBIoT}[NB-IoT]{Narrowband Internet of Things}
\acrodef{NN}{Neural Network}
\acrodef{NSGAII}[NSGA-II]{Non-dominated Sorting Genetic Algorithm II}
\acrodef{PPK}{Power Profiler Kit}
\acrodef{PPKII}[PPK2]{Power Profiler Kit II}
\acrodef{PTQ}{Post-Training Quantization}
\acrodef{PULP}{Parallel Ultra-Low Power}
\acrodef{QAT}{Quantization-Aware Training}
\acrodef{RAM}{Random Access Memory}
\acrodef{RF}{Random Forest}
\acrodef{RISCV}[RISC-V]{Reduced Instruction Set Computer Five}
\acrodef{RMSE}{Root Mean Squared Error}
\acrodef{RNN}{Recurrent Neural Network}
\acrodef{RQ}{Research Question}
\acrodef{SLA}{Standard Latch Array}
\acrodef{SLFNN}{Single-Layer Feedforward Neural Network}
\acrodef{SNN}{Spiking Neural Network}
\acrodef{SRAM}{Static Random Access Memory}
\acrodef{STDC}{Standard Digital Computing}
\acrodef{SVM}{Support Vector Machine}
\acrodef{TinyFL}{Tiny Federated Learning}
\acrodef{TinyML}{Tiny Machine Learning}
\acrodef{UAV}{Unmanned Aerial Vehicle}
\acrodef{UE}{User Equipment}
\acrodef{WFL}{Wireless Federated Learning}

\theoremstyle{thmstyleone}%
\theoremstyle{thmstyletwo}%

\theoremstyle{thmstylethree}%

\begin{document}

\title{Energy Measurement and Optimization of TinyML in Resource-Constrained Embedded Systems: A Systematic Mapping Study}

%%=============================================================%%
%% GivenName	-> \fnm{Joergen W.}
%% Particle	-> \spfx{van der} -> surname prefix
%% FamilyName	-> \sur{Ploeg}
%% Suffix	-> \sfx{IV}
%% \author*[1,2]{\fnm{Joergen W.} \spfx{van der} \sur{Ploeg} 
%%  \sfx{IV}}\email{iauthor@gmail.com}
%%=============================================================%%

\author*[1]{\fnm{Clariele} \sur{Almeida}}
\email{clariele.almeida@ufrpe.br}

\author[1]{\fnm{Renan} \sur{Beserra}}

\author[1]{\fnm{Rafael} \sur{Moura}}

\author[1]{\fnm{Danilo} \sur{Araújo}}

\author[3]{\fnm{Bruno} \sur{Nogueira}}

\author[2]{\fnm{Sérgio} \sur{Chevtchenko}}

\author*[1]{\fnm{Ermeson} \sur{Andrade}}
\email{ermeson.andrade@ufrpe.br}

\affil*[1]{%
  \orgname{Federal Rural University of Pernambuco (UFRPE)},
  \orgaddress{%
    \city{Recife},
    \state{Pernambuco},
    \country{Brazil}
  }
}

\affil[2]{%
  \orgname{Western Sydney University},
  \orgaddress{%
    \city{Sydney},
    \state{New South Wales},
    \country{Australia}
  }
}

\affil[3]{%
  \orgname{Federal University of Alagoas (UFAL)},
  \orgaddress{%
    \city{Maceió},
    \state{Alagoas},
    \country{Brazil}
  }
}

%%==================================%%
%% Sample for unstructured abstract %%
%%==================================%%

\abstract{The rapid expansion of the \ac{IoT} has increased the volume of data generated at the network edge, where many devices operate under severe memory, processing, and energy constraints. In this context, \ac{TinyML} enables \ac{ML} tasks to be executed on microcontrollers, smart sensors, and other resource-constrained platforms. However, the literature remains heterogeneous regarding the algorithms evaluated, energy-measurement methods, reported metrics, influencing factors, and optimization strategies. This study presents a systematic mapping of energy-aware \ac{TinyML} research. It identifies the main models and learning approaches, classifies energy-evaluation methodologies and metrics, examines the technical factors affecting energy consumption, and organizes the reported optimization techniques into a structured taxonomy. The findings highlight the frequent use of convolutional and lightweight architectures, direct measurements and analytical estimations, and factors related to model architecture, hardware, communication, numerical precision, and input characteristics. The optimization strategies were grouped into model compression, hardware-aware design, system and communication optimization, and data and processing optimization. The results also reveal limited standardization in experimental platforms, measurement procedures, and reporting practices, limiting reproducibility and direct comparison across studies. This mapping provides an updated overview of the field and guidance for developing and evaluating more energy-efficient \ac{TinyML} solutions.
}

\keywords{TinyML; energy consumption; energy efficiency; resource-constrained devices; systematic mapping study}

%%\pacs[JEL Classification]{D8, H51}

%%\pacs[MSC Classification]{35A01, 65L10, 65L12, 65L20, 65L70}

\maketitle

\section{Introduction}

\ac{TinyML} refers to the application of \ac{ML} algorithms on embedded devices with limited resources, such as microcontrollers, smart sensors, and battery-powered \ac{IoT} devices \cite{abadade2023comprehensive}. This paradigm enables local data processing, which is essential in applications that require low latency, reduced bandwidth consumption, and enhanced privacy. The main advantage of \ac{TinyML} is that it allows embedded intelligence in environments where continuous access to power or connectivity is not available, making possible solutions such as environmental monitoring, industrial sensors, and wearable devices \cite{zeadally2020design}.

Although \ac{TinyML} and conventional \ac{ML} share theoretical foundations, they differ significantly in terms of scale and computational requirements. Traditional \ac{ML} is typically executed on powerful servers or cloud infrastructure, using complex models and processing large volumes of data \cite{oufettoul2024tinyml}, whereas \ac{TinyML} must be highly optimized to operate efficiently on devices with severe limitations in processing power, memory, and energy \cite{oufettoul2024tinyml}. This contrast makes energy efficiency a central challenge in the development of \ac{TinyML}-based solutions. The use of models that do not consider these constraints can compromise hardware longevity and limit applications in remote or hard-to-reach locations \cite{njor2024holistic}.

The rapid growth of \ac{IoT} further intensifies this scenario. The number of connected devices has increased significantly, generating large volumes of real-time data \cite{cruz2018gddiot}. 
Cloud computing alone is not sufficient to handle this demand efficiently. Studies show that data transmission consumes orders of magnitude more energy than local processing. Hussein et al. \cite{hussein2022distributed} demonstrated that transmitting a 1 KB message over 100 meters consumes energy equivalent to nearly three million microprocessor instructions. Consequently, minimizing data transmission directly translates to substantial energy savings \cite{akyildiz2002wireless}. To address this, edge computing has emerged as a paradigm shift, enabling data processing near the source \cite{cao2020overview}. Beyond energy efficiency, edge computing reduces latency and bandwidth usage while enhancing security and privacy, as highlighted in \cite{coimbra2023previsao}. Nevertheless, deploying \ac{ML} models on edge devices introduces new challenges, particularly due to stringent energy and computational constraints. Thus, as emphasized by \cite{feng2022computation}, developing energy-efficient algorithms is important for viable \ac{TinyML} applications in resource-limited environments.

Despite the growing interest in machine learning solutions for resource-limited devices, the literature still lacks studies that carry out a systematic mapping focused exclusively on energy consumption in \ac{TinyML} algorithms. Existing reviews primarily address \ac{TinyML} from a broad perspective, emphasizing applications, tools, and general challenges, while treating energy efficiency only as a secondary aspect. Works such as those by Capogrosso et al. \cite{capogrosso2024machine} and Abadade et al. \cite{abadade2023comprehensive} provide comprehensive overviews of \ac{TinyML} without delving into energy consumption analysis. Similarly, Kallimani et al. \cite{kallimani2024tinyml} discuss tools, metrics, and future perspectives related to \ac{TinyML}, including energy consumption as one of the evaluation criteria, but do not structure a systematic categorization of energy measurement or optimization techniques. In related contexts, Silva and Prazeres \cite{da2025tiny} and Arif and Rashid \cite{arif2025literature} explore \ac{TinyFL} and \ac{EdgeML}, respectively, incorporating references to energy efficiency. Nevertheless, none of these studies offers a comprehensive and systematic analysis of the approaches, metrics, technical factors, and optimization strategies specifically associated with energy consumption in \ac{TinyML} algorithms, which hinders direct comparison between approaches and the consolidation of best practices.

In light of this gap, this work presents a systematic mapping of the literature on energy consumption in \ac{TinyML} solutions deployed on resource-constrained devices. The study examines \ac{TinyML} algorithms used in energy-aware deployments, the metrics and methodologies adopted to measure and evaluate energy consumption, the technical factors that influence these measurements on edge devices, and the optimization techniques proposed to reduce energy consumption based on empirical evidence. 
The analysis focuses primarily on the inference and deployment phases of  \ac{TinyML} models on resource-constrained devices, while also including studies that explicitly evaluate the energy impact of on-device learning, retraining, or federated learning.
The analysis covers studies published between 2020 and 2025 and provides a structured synthesis to support future research and the design of energy-efficient \ac{TinyML} solutions.
More specifically, this study makes the following contributions:
\begin{enumerate}
    \item Maps \ac{TinyML} algorithms evaluated under energy constraints on resource-limited devices;
    \item Classifies energy-measurement approaches and reported metrics;
    \item Identifies technical factors that affect energy consumption during TinyML deployment and execution; and
    \item Organizes optimization strategies into a taxonomy covering model compression, hardware-aware design, system and communication aspects, and data-processing techniques.
\end{enumerate}

The remainder of this article is structured as follows: Section \ref{sec2} reviews existing studies on energy consumption in \ac{TinyML} algorithms. Section \ref{mapping} describes the protocol adopted to conduct the research. Section \ref{results} presents the results obtained for the defined research questions. Section \ref{open} discusses the unresolved challenges. Section \ref{lim} outlines the limitations of the Systematic Mapping, and finally, Section \ref{con} presents the conclusions and suggests possible directions for future research.

\section{Related Work} 
\label{sec2}

This section discusses recent literature reviews on \ac{TinyML} and related paradigms, including \ac{TinyFL}, \ac{EdgeML}, and \ac{EdgeGenAI}. These studies cover different aspects of the field, such as applications, hardware platforms, software frameworks, learning techniques, deployment strategies, and optimization methods. Although energy consumption appears in several of these reviews, it is usually treated as one evaluation criterion among others rather than as the central focus of analysis. As a result, the way energy consumption is measured, analyzed, and optimized remains only partially covered in the existing literature. 
%Table~\ref{tab:related_work} summarizes the main characteristics of these works and contrasts them with the focus of the present study.

The study by Capogrosso et al.~\cite{capogrosso2024machine} presents an updated systematic review of the \ac{TinyML} field, with an emphasis on learning techniques designed for resource-constrained devices such as \acp{MCU} and smart sensors. It summarizes the most widely used hardware platforms, model optimization strategies, and current applications, as well as the major challenges in this domain, but without providing a systematic or in-depth analysis of how energy is measured, which techniques are used to optimize it, or which technical factors most strongly influence it.

Silva and Prazeres \cite{da2025tiny} present a systematic review on the \ac{TinyFL} paradigm, which integrates \ac{FL} and \ac{TinyML} to enable local model training on extremely resource-limited devices, such as microcontrollers. Their study highlights the versatility of \ac{TinyFL}, with applications in areas such as the \ac{IoT}, the \ac{IoMT}, and \acp{UAV}. It also analyzes aggregation algorithms, compression techniques, and frameworks such as TensorFlow Lite Micro, and explores strategies for adapting \ac{FL} to low-power devices such as the ESP32, Arduino Nano \ac{BLE}, and Raspberry Pi.

The review by Dutta and Bharali \cite{dutta2021tinyml} investigates the integration between \ac{TinyML} and \ac{IoT} devices, highlighting its potential for efficient operation in resource-constrained environments. The authors emphasize that applying \ac{TinyML} on \acp{MCU} enables local data processing, reducing dependence on often overloaded cloud services and minimizing the energy consumption associated with continuous data transmission. The study presents the historical development of the field and synthesizes the main benefits of \ac{TinyML}, with emphasis on metrics such as latency, energy consumption, reliability, privacy, and cost. The review also discusses the importance of hardware-software co-design to overcome \ac{MCU} limitations, analyzing compression techniques such as pruning, quantization, knowledge distillation, and deep compression. Additionally, it presents popular frameworks such as TensorFlow Lite Micro and \ac{CMSISNN}, along with real-world applications in areas such as agriculture, smart cities, and healthcare systems, reinforcing the feasibility of \ac{TinyML} in battery-powered and low-connectivity devices.

The work developed by {Arif and Rashid}~\cite{arif2025literature} reviews model conversion techniques, inference mechanisms, and learning strategies in the context of \ac{EdgeML} and \ac{TinyML}, focusing on implementation in resource-constrained edge devices. The study explores state-of-the-art model conversion techniques used in \ac{EdgeML} and analyzes their impact on performance, efficiency, and the feasibility of deploying machine learning models on constrained devices. It also investigates modern inference mechanisms, comparing latency, energy consumption, and model accuracy during local execution. Moreover, the article discusses how different learning strategies, including continual learning, federated learning, and on-device learning, affect the performance and efficiency of models deployed in edge computing environments. It also presents the latest advances in hardware, software frameworks, and sensors specifically designed for \ac{TinyML} applications.

Abadade et al.~\cite{abadade2023comprehensive} provide a comprehensive review of \ac{TinyML} applications and developments, proposing a taxonomy of the main techniques used, organized by domains such as healthcare, agriculture, environment, and anomaly detection. The study compares \ac{TinyML} with traditional approaches such as cloud, fog, and edge computing, emphasizing the advantages of local processing, particularly in terms of latency and energy consumption. Although energy efficiency is not the central focus, the authors note that energy savings result mainly from reduced data transmission and processing needs. Strategies such as the use of low-power components, low-voltage operation, and sleep modes are also mentioned. The study includes a comparative table of \ac{TinyML} devices, detailing hardware specifications, energy consumption, connectivity, sensors, and costs.

Kallimani et al. \cite{kallimani2024tinyml} present a comprehensive review of the \ac{TinyML} field, covering its foundations, tools, applications, challenges, and future directions. The work describes key frameworks and hardware platforms compatible with \ac{TinyML}, such as Scikit-learn, TensorFlow, and PyTorch, and highlights the main evaluation metrics used for \ac{TinyML} models, including accuracy, energy consumption, latency, and memory requirements. Giorgetti and Pau \cite{giorgetti2025transitioning} present a review focused on the transition from traditional \ac{TinyML} solutions to \ac{EdgeGenAI}. The paper highlights how the increasing availability of on-device accelerators and low-power \ac{AI} cores is enabling more complex workloads at the edge, including lightweight generative models. %The authors survey existing TinyML approaches and compare them with emerging Edge GenAI pipelines in terms of model size, memory requirements, and energy consumption.

 More recent domain-specific surveys have expanded the discussion on \ac{TinyML} applications in human-centered scenarios. Lamaakal et al. \cite{Lamaakal25} presented a comprehensive survey on \ac{TinyML} for \ac{HBA}, covering \ac{TinyML} concepts, hardware and software foundations, model compression techniques, benchmark datasets, and applications such as human activity recognition, healthcare, transportation, education, workplace monitoring, public safety, and sentiment analysis. Essahraui and Lamaakal \cite{essahraui2026comprehensive} surveyed \ac{TinyML}-based biometric recognition for \ac{IoT} edge devices, addressing physiological and behavioral biometric modalities, including fingerprint, face, iris, gait, voice, and keystroke dynamics, as well as application domains such as wearable devices, smart home and \ac{IoT} security, healthcare monitoring, access control systems, and surveillance.

Table~\ref{tab:related_work} summarizes how the present study differs from previous reviews. Existing studies either provide broad overviews of \ac{TinyML}, address related paradigms such as \ac{TinyFL}, \ac{EdgeML}, and \ac{EdgeGenAI}, or focus on specific application domains such as \ac{HBA} and biometric recognition. Although several of these works mention energy efficiency, they usually treat it as one evaluation criterion among others, rather than as the central object of analysis. In contrast, this study focuses specifically on the energy consumption of \ac{TinyML} algorithms implemented on resource-constrained devices. It applies clearly defined research questions and selection criteria to classify \ac{TinyML} algorithms, energy-measurement methodologies, reported metrics, technical factors affecting energy consumption, and optimization strategies. By consolidating these dimensions, this work provides a structured mapping of energy-aware \ac{TinyML} research and identifies gaps related to measurement standardization, hardware-dependent evaluation, and energy-performance trade-offs.

\begin{table*}[!ht]
\centering
\caption{Summary of related \ac{TinyML} reviews.}
\label{tab:related_work}
\scriptsize
\setlength{\tabcolsep}{2.5pt}
\renewcommand{\arraystretch}{1.0}
\begin{tabularx}{\textwidth}{p{0.7cm} p{2.2cm} p{2.7cm} X}
\toprule
\textbf{Year} & \textbf{Study} & \textbf{Focus} & \textbf{Main characteristics} \\
\midrule
2024 & Capogrosso et al.~\cite{capogrosso2024machine}
& \ac{TinyML} learning techniques
& Learning techniques, hardware platforms, optimization strategies, applications, and challenges. \\

2025 & Silva and Prazeres~\cite{da2025tiny}
& Tiny Federated Learning
& \ac{TinyFL} on constrained devices, aggregation algorithms, compression methods, frameworks, and low-power platforms. \\

2021 & Dutta and Bharali~\cite{dutta2021tinyml}
& \ac{TinyML} for \ac{IoT}
& Local processing, reduced cloud dependence, latency, privacy, cost, energy savings, and model compression. \\

2025 & {Arif and Rashid}~\cite{arif2025literature}
& \ac{EdgeML} and \ac{TinyML}
& Model conversion, inference mechanisms, learning strategies, latency, energy consumption, and accuracy. \\

2023 & Abadade et al.~\cite{abadade2023comprehensive}
& \ac{TinyML} applications
& Applications in healthcare, agriculture, environment, and anomaly detection; energy efficiency as a secondary aspect. \\

2024 & Kallimani et al.~\cite{kallimani2024tinyml}
& \ac{TinyML} tools and challenges
& Foundations, frameworks, platforms, applications, challenges, and metrics such as accuracy, latency, memory, and energy. \\

2025 & Giorgetti and Pau~\cite{giorgetti2025transitioning}
& \ac{EdgeGenAI}
& Transition from \ac{TinyML} to \ac{EdgeGenAI}, including accelerators, low-power \ac{AI} cores, memory limits, and energy constraints. \\

2025 & Lamaakal et al.~\cite{Lamaakal25}
& \ac{TinyML} for \ac{HBA}
& Human-centered \ac{TinyML} applications, datasets, compression, latency, memory, and energy consumption. \\

2026 & Essahraui and Lamaakal~\cite{essahraui2026comprehensive}
& \ac{TinyML} biometrics
& \ac{TinyML}-based biometric recognition on \ac{IoT} edge devices, covering physiological and behavioral modalities. \\

\textbf{2026} & \textbf{This study}
& \textbf{Energy consumption in \ac{TinyML}}
& \textbf{\ac{TinyML} algorithms, energy-measurement methods, metrics, technical factors, optimization techniques, and research gaps.} \\
\bottomrule
\end{tabularx}
\end{table*}

\section{Systematic Mapping Study Process}
\label{mapping}

\begin{figure}[h!]
    \centering    \includegraphics[width=0.9\linewidth]{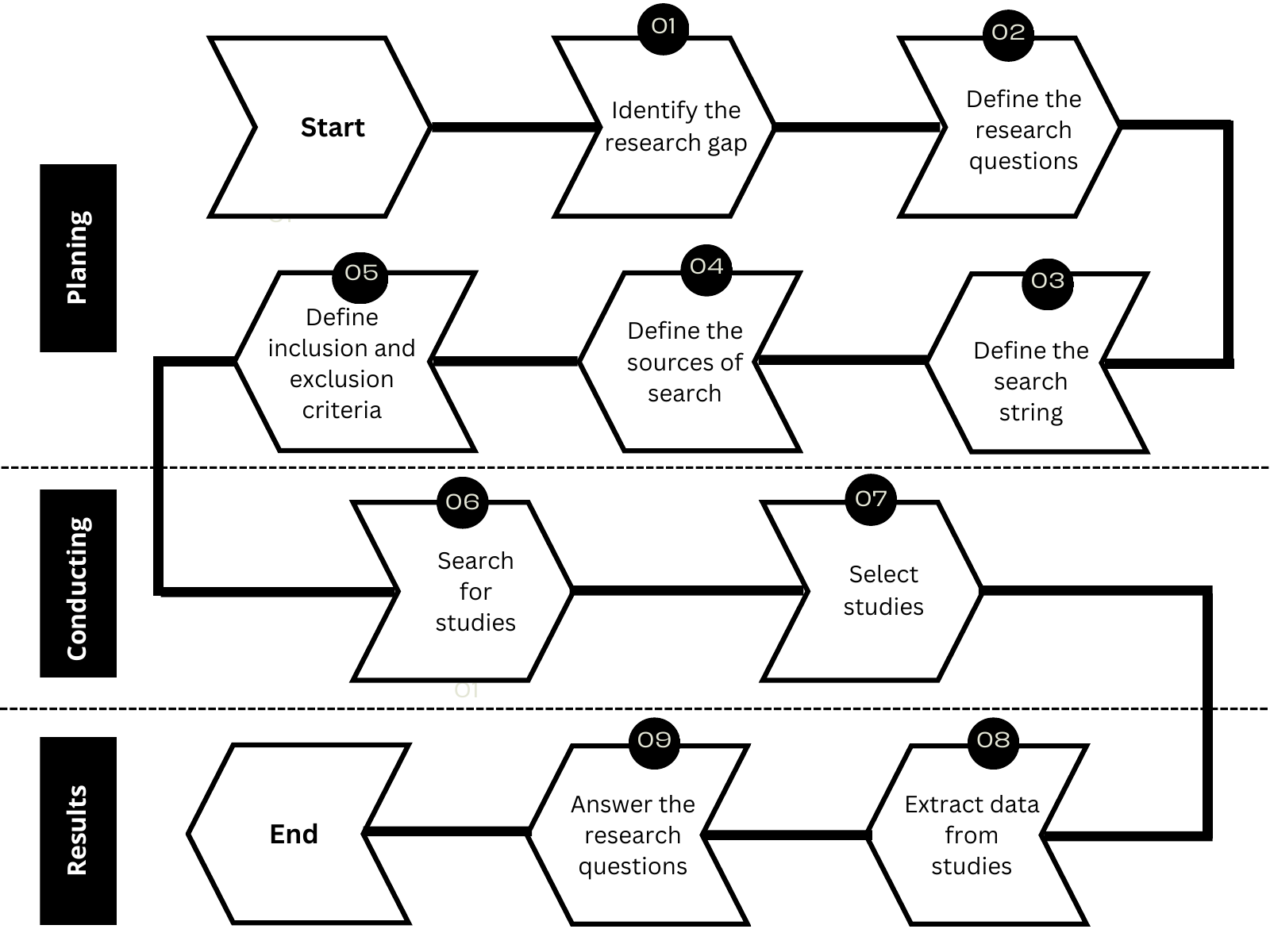}
    \caption{Steps of the adopted methodology.}
    \label{fig:metodology}
\end{figure}

This section presents the steps of the protocol adopted to conduct the Systematic Mapping of the literature, with the aim of identifying studies related to energy consumption in \ac{TinyML} algorithms implemented on resource-constrained devices. For this purpose, we followed the methodology proposed by Petersen et al. \cite{petersen2008systematic}.\footnote{The complete replication package is publicly available at \url{https://zenodo.org/records/23348211}.} Figure~\ref{fig:metodology} illustrates the steps of the adopted process.

\textbf{Step 1: Identify the research gap.} Energy consumption in \ac{TinyML} algorithms has become an increasing concern, especially due to the hardware limitations and limited energy autonomy of embedded devices \cite{elhanashi2024advancements}. With the advancement of edge computing and the need for local data processing, it is essential to understand the factors that influence energy consumption and the strategies for optimizing it \cite{immonen2022tiny}. However, challenges remain in standardizing metrics, measurement methodologies, and efficient techniques for reducing energy consumption in \ac{TinyML} models.

In light of this scenario, this systematic mapping aims to investigate the approaches and techniques used to evaluate and optimize energy consumption in \ac{TinyML} algorithms deployed on resource-constrained devices. The goal is to identify the most commonly used \ac{TinyML} algorithms, the metrics and methodologies employed to measure energy consumption, the factors that impact these measurements, and the techniques proposed to optimize energy efficiency.

By consolidating and analyzing the available studies in the literature, this mapping contributes to a better understanding of the state-of-the-art in energy efficiency for \ac{TinyML}, providing an updated overview of the area and highlighting trends, challenges, and opportunities for future research.

\textbf{Step 2: Define the \acp{RQ}.} This Systematic Mapping seeks to explore the current state of research on energy consumption in \ac{TinyML} algorithms implemented on resource-constrained devices. The following research questions were defined:

\begin{itemize}
    \item \textbf{RQ1} - What are the main \ac{TinyML} algorithms applied to resource-constrained devices with a focus on energy consumption?
    \item \textbf{RQ2} - What metrics and methodologies are used to measure and evaluate energy consumption in \ac{TinyML} algorithms on resource-limited devices? 
    \item \textbf{RQ3} - What technical factors influence energy consumption measurements in \ac{TinyML} algorithms deployed on edge computing devices? 
    \item \textbf{RQ4} - What techniques have been proposed to optimize the energy consumption of \ac{TinyML} algorithms on edge computing devices based on measurements? 
\end{itemize}

To answer \textbf{RQ1}, we extracted the \ac{TinyML} algorithms adopted in each selected study and classified them into higher-level model families (e.g., \acp{CNN} and variants, lightweight architectures, and classical machine learning methods). We then analyzed the frequency with which each algorithm appeared and identified the application domains in which they were used. In addition, we recorded whether optimization techniques (e.g., quantization, pruning, and knowledge distillation) were applied. The results were summarized using descriptive statistics, particularly frequency counts, and visualized through bar charts and tables.

To answer \textbf{RQ2}, we identified the methodologies used to measure energy consumption and grouped them into predefined categories: direct measurement, analytical estimation, inference-time-based estimation, and simulation. We also extracted the metrics reported in each study (e.g., energy per inference, inference time, and power consumption) and summarized them in tabular form. When possible, we related the choice of metric to the objective of the study.

To answer \textbf{RQ3}, we extracted the technical factors reported as influencing energy consumption (e.g., processor type, memory hierarchy, model architecture, input size, and communication protocol). The frequency of each factor was counted and presented using bar charts, highlighting the factors most commonly reported in the literature.

Finally, to answer \textbf{RQ4}, we extracted the optimization techniques employed in the selected studies and classified them according to a four-category taxonomy: Model Compression, Hardware-Aware Design, System \& Communication, and Data \& Processing. The key techniques in each category were listed together with the supporting studies, and their prevalence was summarized using descriptive statistics and visual representations.

\textbf{Step 3: Define the search string.} To retrieve the primary studies, an effective search string was developed for automated queries across the major digital libraries. The definition of the search terms was based on the authors’ knowledge of the field and refined using exploratory searches. Initially, key terms were identified from common terminology in the literature related to \ac{TinyML}, energy consumption, and edge computing. These terms were then iteratively combined using Boolean operators (AND, OR) to generate alternative formulations of the search string.

To validate the search strategy, pilot searches were conducted in Google Scholar to assess whether alternative query formulations were able to retrieve known relevant studies. During this process, terms were added, removed, and reordered to improve the coverage of relevant studies while limiting the retrieval of clearly unrelated publications. The final string was intentionally centered on energy consumption and its measurement in resource-constrained edge environments, since these dimensions define the scope of this mapping. The objective was not to retrieve all studies proposing or evaluating TinyML algorithms, but specifically those that provide explicit evidence regarding energy consumption, power usage, or energy-related deployment costs. Many recent machine learning algorithms have not yet been evaluated from an energy-consumption perspective.  Consequently, including their names would substantially increase the number of studies outside the scope without necessarily improving the retrieval of relevant energy-oriented evidence. Moreover, enumerating specific algorithms could bias the search toward predefined model families and exclude emerging or application-specific approaches. Therefore, ``TinyML'' was adopted as the main technological descriptor, while the remaining terms were used to delimit the energy-measurement and resource-constrained deployment context. The final version was selected after consistently retrieving a predefined set of known relevant publications and maintaining alignment with the research questions. The resulting search string is as follows:
("\ac{TinyML}") AND ("energy consumption") AND ("edge computing") AND ("measurement") AND ("resource-constrained").

%This string aims to capture studies addressing the energy consumption of TinyML algorithms on resource-limited devices, considering measurement metrics and the edge computing context.

\textbf{Step 4: Define the sources of search.}
After defining the search string, five digital libraries were selected as the primary sources for the study. The chosen libraries were:
ACM Digital Library\footnote{http://portal.acm.org},
IEEE Digital Library\footnote{http://ieeexplore.ieee.org},
ScienceDirect\footnote{http://www.sciencedirect.com},
Scopus\footnote{http://www.scopus.com}, and
Springer Link\footnote{http://link.springer.com}. These digital libraries were selected because they are widely recognized and commonly used as primary sources in systematic reviews and mapping studies in this research area, ensuring broad coverage of the relevant literature.
{The search was applied to the metadata fields (titles, abstracts, and indexing terms) in IEEE Xplore and Scopus, and to the full text in ACM Digital Library, ScienceDirect, and Springer Link. The verbatim strings, search fields, and filters used in each library are available in the replication package.}

\textbf{Step 5: Define inclusion and exclusion criteria.} Given the large number of studies returned by the databases using the defined search string, and considering that conducting a systematic mapping with an excessively large number of articles would be unfeasible, many studies were excluded for being irrelevant to this research. Thus, inclusion and exclusion criteria were established.

To be included, studies needed to address both \ac{TinyML} and energy consumption analysis simultaneously. The exclusion criteria were: (i) studies published outside the period from 2020 to 2025, (ii) surveys, books, and proceedings volumes/editorial records, (iii) duplicate studies, (iv) articles presenting unclear results or findings and (v) studies not focused on \ac{TinyML} and/or energy consumption analysis. It should be noted that the publication period (2020--2025) was applied directly as a search filter in each digital library whenever this option was available. Therefore, studies published outside this period were not retrieved {by the main search} and, consequently, were not excluded during the screening process. {The only exceptions were six records from the supplementary IEEE Xplore search, which were published after the main search date and were excluded under criterion (i).}

\textbf{Step 6: Search for studies.} 
At this stage, the search string was applied to the selected digital libraries to identify relevant articles for the study. The searches were conducted on January 12, 2026, resulting in the retrieval of 209 studies.
{Because the main search returned only two records from IEEE Xplore, a supplementary search was conducted on this library on October 2, 2026, to verify its coverage, using the same string without the term ``measurement'' (see Section~\ref{lim}). This search returned 19 records.}

\textbf{Step 7: Select the studies.} After retrieving the articles using the search string, a preliminary analysis was conducted by applying the previously defined inclusion and exclusion criteria. Initially, two authors independently screened the titles, abstracts, and keywords of the 209 retrieved studies, identifying 35 potentially relevant articles for full-text assessment. The same reviewers then independently examined the full text of these 35 studies. In both phases, disagreements or uncertainties were resolved through direct discussion until consensus was reached, and third-party review was not required. 
%Following the full-text assessment, 23 studies satisfied all inclusion criteria and were therefore selected for data extraction and in-depth analysis. 
{Following the full-text assessment, 23 studies from the main search satisfied all inclusion criteria and were therefore selected for data extraction and in-depth analysis. The 19 records from the supplementary IEEE Xplore search were screened by the same two reviewers using the same criteria; ten of them were assessed in full text, and two additional studies were included, resulting in a final set of 25 primary studies.}
These studies demonstrated alignment with the research questions and addressed the energy consumption of TinyML algorithms implemented on resource-constrained devices. 

\textbf{Step 8: Extract data from studies.} In this phase, relevant data were extracted through a detailed reading of the selected articles. The information collected was directed toward answering the research questions, enabling a systematic analysis of the main findings. To ensure the validity and consistency of the data extraction process, a collaborative multi-reviewer approach was adopted. Each selected study was analyzed by at least two authors, who independently extracted information based on predefined variables. The results were then compared, and any
discrepancies were discussed until a consensus was reached regarding the extracted data. This strategy helped minimize individual biases and increase the reliability of the collected information.

\textbf{Step 9: Answer the research questions.} Finally, the data collected in the previous step were organized into charts and tables to facilitate the analysis and formulation of answers to the research questions. Section~\ref{results} provides an overview of the results obtained, along with the responses to the defined questions.

\section{Results and Discussion}
\label{results}

This section presents and analyzes the main findings of this systematic mapping study on energy consumption in \ac{TinyML} algorithms deployed on resource-constrained devices. The selected studies were organized to provide a clear view of trends, challenges, and advancements in the field. Initially, the characteristics of the included articles are explored, such as the temporal distribution of publications, consulted databases, and types of publications. Subsequently, we discuss the most frequently used \ac{TinyML} algorithms focused on energy efficiency, the metrics and methodologies applied to measure energy consumption, the technical factors that influence these measurements, and the optimization strategies identified in the literature.
The analysis of the results aims to answer the research questions, offering a structured understanding of existing approaches, as well as the gaps and challenges that still need to be addressed. This section thus contributes to a deeper understanding of the impact of energy consumption in \ac{TinyML} and the possibilities for optimization in this context.

\subsection{Overview of Selected Studies}

Table~\ref{tab:exclusion} presents the number of articles excluded according to each criterion. %From the {209} studies initially identified through the search string in the consulted databases, 74 were excluded for being proceedings volumes or editorial records, and 46 for not addressing the core topic of this work: energy consumption analysis in \ac{TinyML} algorithms. In addition, 52 studies were excluded for being surveys and another 12 for being books or book chapters. One article was removed for being duplicated, and another for presenting unclear results.
{From the 228 records identified (209 from the main search and 19 from the supplementary IEEE Xplore search), 74 were excluded for being proceedings volumes or editorial records, and 54 for not addressing the core topic of this work: energy consumption analysis in TinyML algorithms. In addition, 53 studies were excluded for being surveys and another 12 for being books or book chapters. Six records from the supplementary search were published in 2026, outside the 2020--2025 period. Three articles were removed for being duplicated (two of them corresponding to studies already retrieved by the main search), and another for presenting unclear results.}

\begin{table*}[h!]
\centering
\scriptsize % Tamanho reduzido de fonte
\caption{{Studies Exclusion Criteria.}}
\label{tab:exclusion}
\begin{tabular}{lc}
\hline
Exclusion Criteria & Articles Excluded  \\ \hline
Proceedings volumes/editorial records       & 74 \\
Articles Missing \ac{TinyML} and/or Energy Consumption Analysis & 54 \\
Articles that are Surveys      & 53 \\
Books or book chapters              & 12 \\
Published outside the 2020--2025 period & 6 \\
Duplicated              & 3 \\
Unclear results              & 1 \\ \hline
\end{tabular}
\end{table*}

\begin{figure}[!h]
    \centering
    \includegraphics[width=0.9\linewidth]{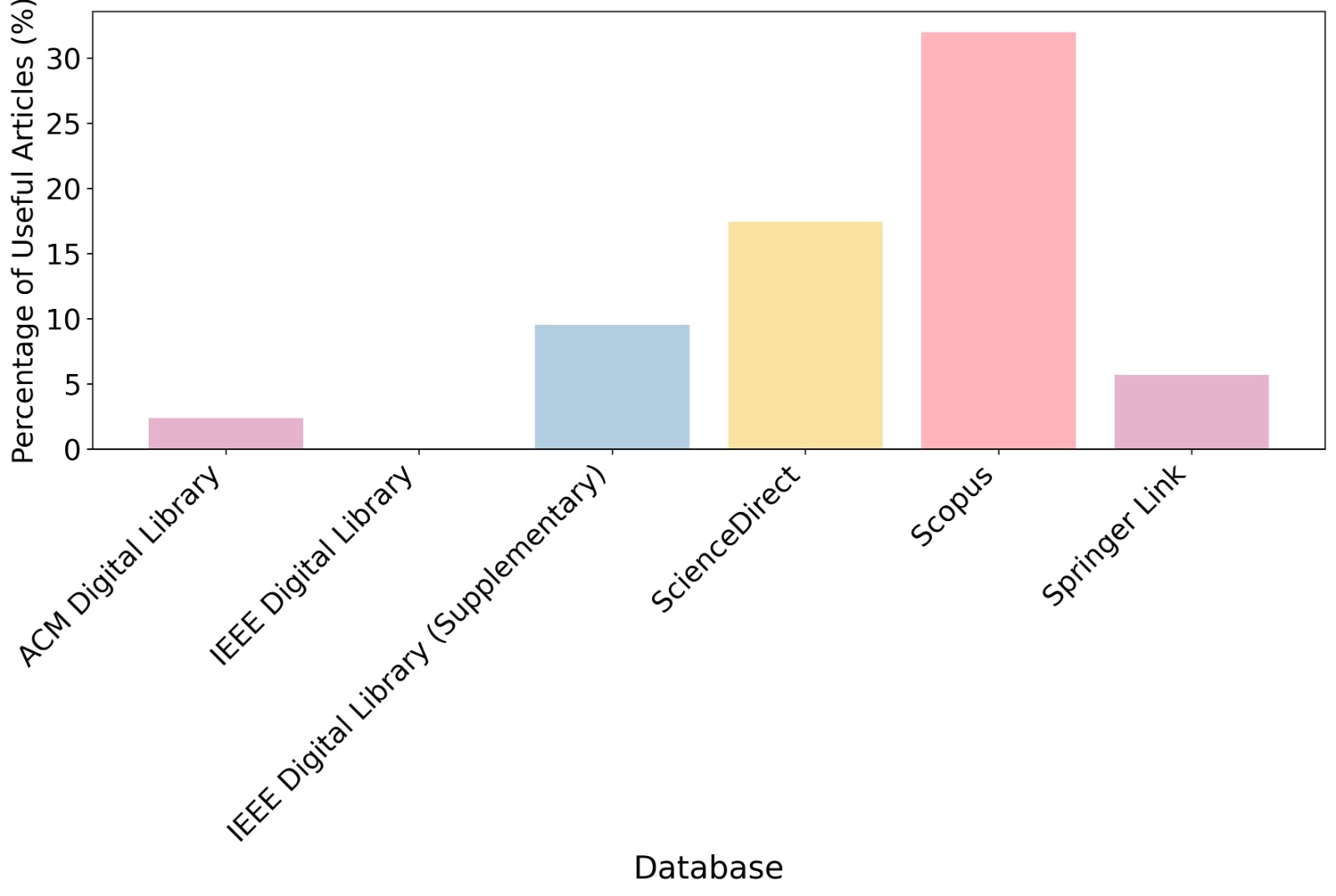}
\caption{Inclusion Rate by Digital Library.}
    \label{fig:database_usage_distribution}
\end{figure}

% Table~\ref{tab:studies} shows the number of studies before and after applying the inclusion and exclusion criteria. After the initial screening and application of the established criteria, 23 articles were selected for in-depth analysis in order to address the research questions. This number represents approximately 11\% of the total identified studies, highlighting the importance of well-defined criteria for selecting relevant work.
% When analyzing the proportion of selected articles, Scopus yielded 8 selected studies, corresponding to an inclusion rate of 32.00\%, the highest among the consulted sources. ScienceDirect contributed 11 selected studies, with an inclusion rate of 17.46\%. SpringerLink and the ACM Digital Library had lower inclusion rates of 5.71\% and 2.38\%, respectively, while none of the studies retrieved from the IEEE Xplore Digital Library met the established selection criteria.These proportions are shown in Figure~\ref{fig:database_usage_distribution}.
{Table~\ref{tab:studies} shows the number of studies before and after applying the inclusion and exclusion criteria. After the initial screening and application of the established criteria, 25 articles were selected for in-depth analysis in order to address the research questions. This number represents approximately 11\% of the total identified records, highlighting the importance of well-defined criteria for selecting relevant work. When analyzing the proportion of selected articles, Scopus yielded 8 selected studies, corresponding to an inclusion rate of 32.00\%, the highest among the consulted sources. ScienceDirect contributed 11 selected studies, with an inclusion rate of 17.46\%. SpringerLink and the ACM Digital Library had lower inclusion rates of 5.71\% and 2.38\%, respectively, while none of the studies retrieved from the IEEE Xplore Digital Library in the main search met the established selection criteria. The supplementary IEEE Xplore search contributed two additional studies, corresponding to an inclusion rate of 10.53\%. The low yield of IEEE Xplore in the main search results from the metadata-only matching of the term ``measurement'' and is discussed in Section~\ref{lim}. These proportions are shown in Figure~\ref{fig:database_usage_distribution}.}

\begin{table}[h!]
\centering
\scriptsize
\caption{Studies before and after the application of inclusion and exclusion criteria.}
\label{tab:studies}
\begin{tabular}{lcc}
\hline
Database & Original & After Criteria \\ \hline
ACM Digital Library & 84 & 2 \\
IEEE Digital Library & 2 & 0 \\
{IEEE Digital Library (supplementary search)\footnotemark[1]} & 19 & 2 \\
ScienceDirect      & 63 & 11 \\
Scopus              & 25 & 8 \\
Springer Link       & 35 & 2 \\ \hline
\end{tabular}
\footnotetext[1]{Search without the term ``measurement''. Two additional records were duplicates of studies already included.}
\end{table}

\begin{figure}[!]
    \centering
    \includegraphics[width=0.9\linewidth]{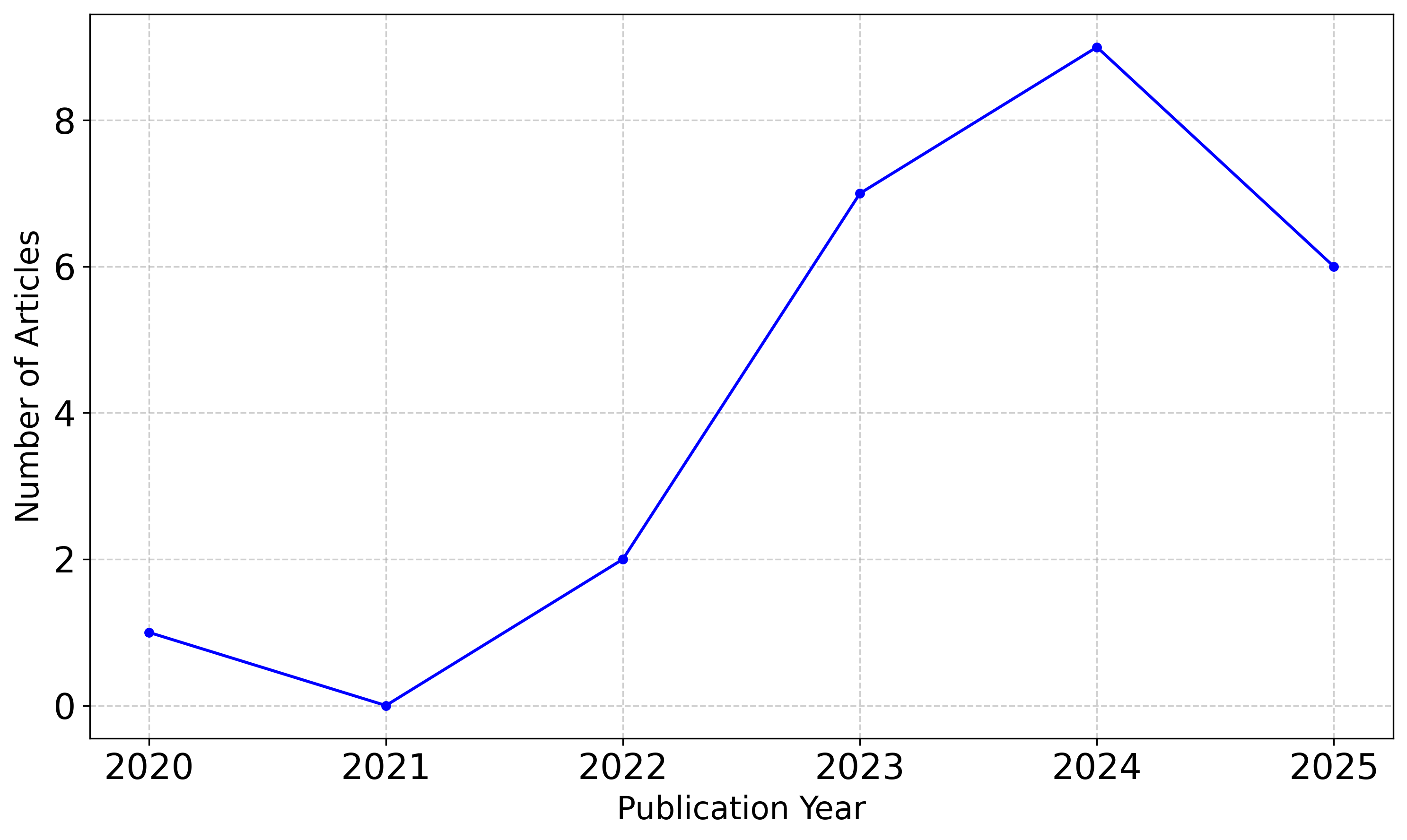}
    \caption{Publication Frequency By Year.}
    \label{fig:publication_frequency_by_year}
\end{figure}

Figure~\ref{fig:publication_frequency_by_year} illustrates the distribution of selected studies by year of publication. Although the search considered articles published between 2020 and 2025, only one study was included for 2020 and {two} for 2022, while no studies met the selection criteria for 2021. A substantial increase in the number of publications was observed from 2023 onward. The highest number of selected studies was identified in 2024, with {nine} publications, followed by seven studies in 2023 and six in 2025. This distribution indicates a growing research interest in energy-aware \ac{TinyML} in recent years, with publication activity remaining consistently high after 2023.

\begin{figure}[!]
    \centering
    \includegraphics[width=0.9\linewidth]{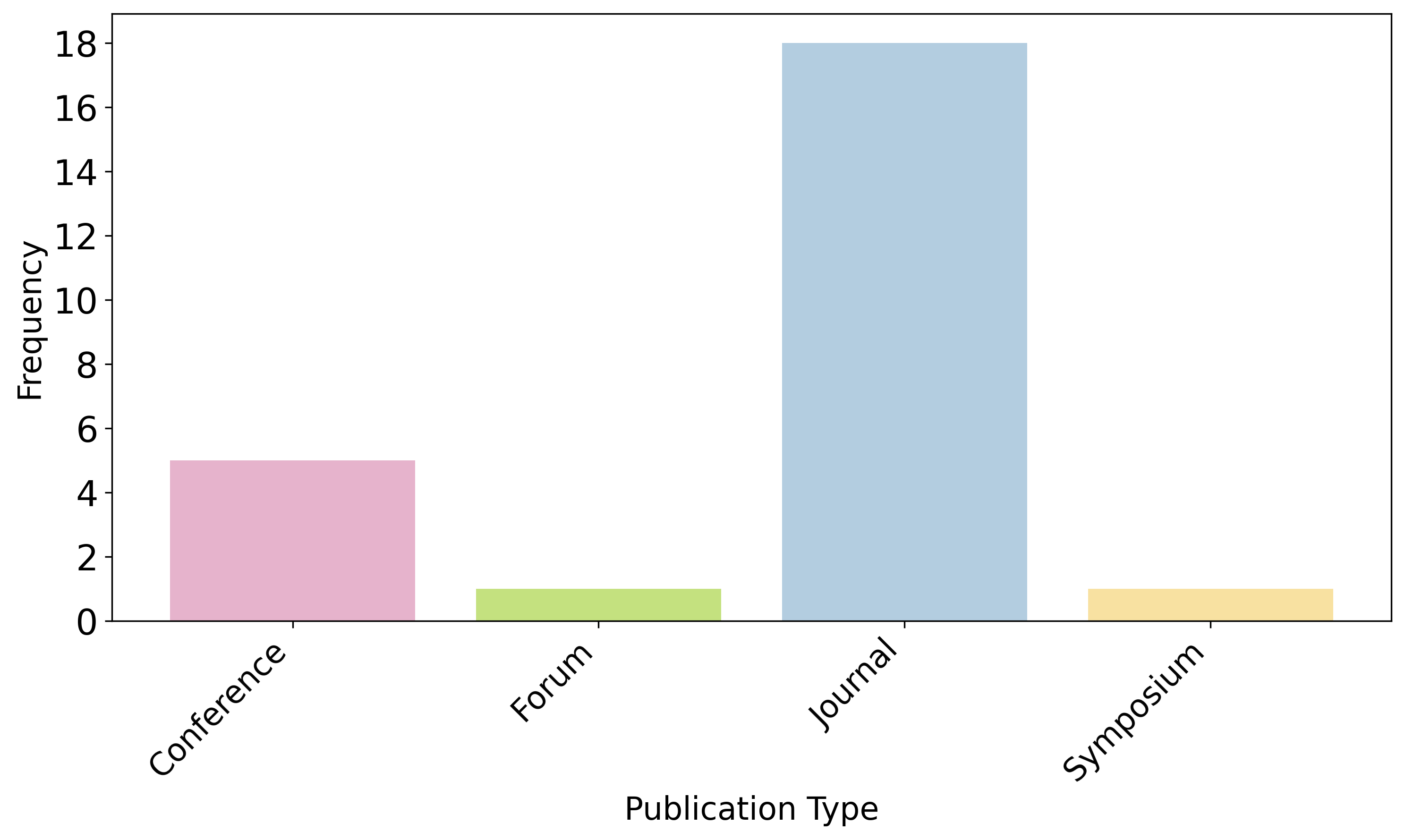}
    \caption{Publication Type Frequency.}
    \label{fig:publication_types_frequency}
\end{figure}

Figure~\ref{fig:publication_types_frequency} presents the distribution of publication types identified in the systematic mapping study. The results show that most analyzed articles were published in journals, accounting for {18} occurrences. Conferences were the second most common type, with 5 articles. This indicates that the topic is also being discussed at academic events, where new findings and advances are often presented before being published in journals. Finally, symposia and forums were each represented by only 1 occurrence.

\begin{figure}[!]
    \centering    \includegraphics[width=0.8\linewidth]{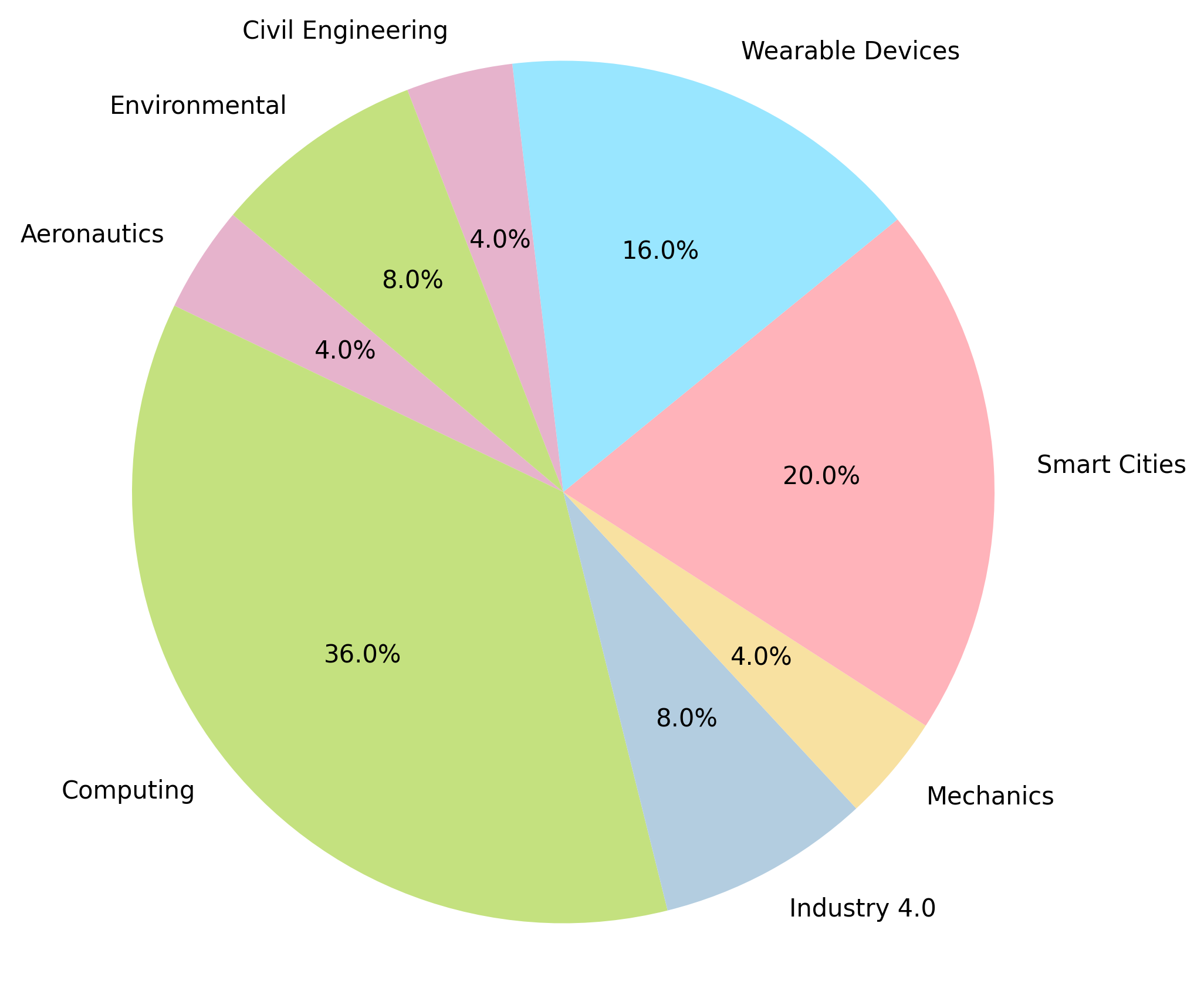}
    \caption{Domain Application Frequency.}
    \label{fig:domain_application_frequency}
\end{figure}

Among the {25} selected studies, the application domains were diverse, including but not limited to: object hardness classification, activity classification for power tools in the construction sector, physical exercise recognition, sign language gesture recognition, detection of attacks exploiting resource constraints (energy and memory) in \ac{IoT} devices, and anomaly detection in urban noise time series. 
These studies were categorized into eight application areas (see Figure~\ref{fig:domain_application_frequency}). 
%Computing was the most frequent domain, accounting for 39.1\% of the articles, followed by Wearable Devices and Smart Cities, each representing 17.4\%. Industry 4.0 accounted for 8.7\%, while Mechanics, Aeronautics, Environmental, and Civil Engineering each represented 4.3\%.
{Computing was the most frequent domain, accounting for 36.0\% of the articles, followed by Smart Cities (20.0\%) and Wearable Devices (16.0\%). Industry 4.0 and Environmental each accounted for 8.0\%, while Mechanics, Aeronautics, and Civil Engineering each represented 4.0\%.}

The distribution of application domains shows that the selected \ac{TinyML} studies are concentrated particularly in Computing, {Smart Cities, and Wearable Devices.} The presence of studies in Industry 4.0, Environmental monitoring, Mechanics, Aeronautics, and Civil Engineering also demonstrates the applicability of energy-aware \ac{TinyML} across heterogeneous operational contexts in which resource efficiency, autonomy, and local processing are relevant deployment requirements.

\subsection{RQ1- What Are the Main \ac{TinyML} Algorithms Applied on Resource-Constrained Devices with a Focus on Energy Consumption?}

\begin{figure}[!]
    \centering  \includegraphics[width=1\linewidth]{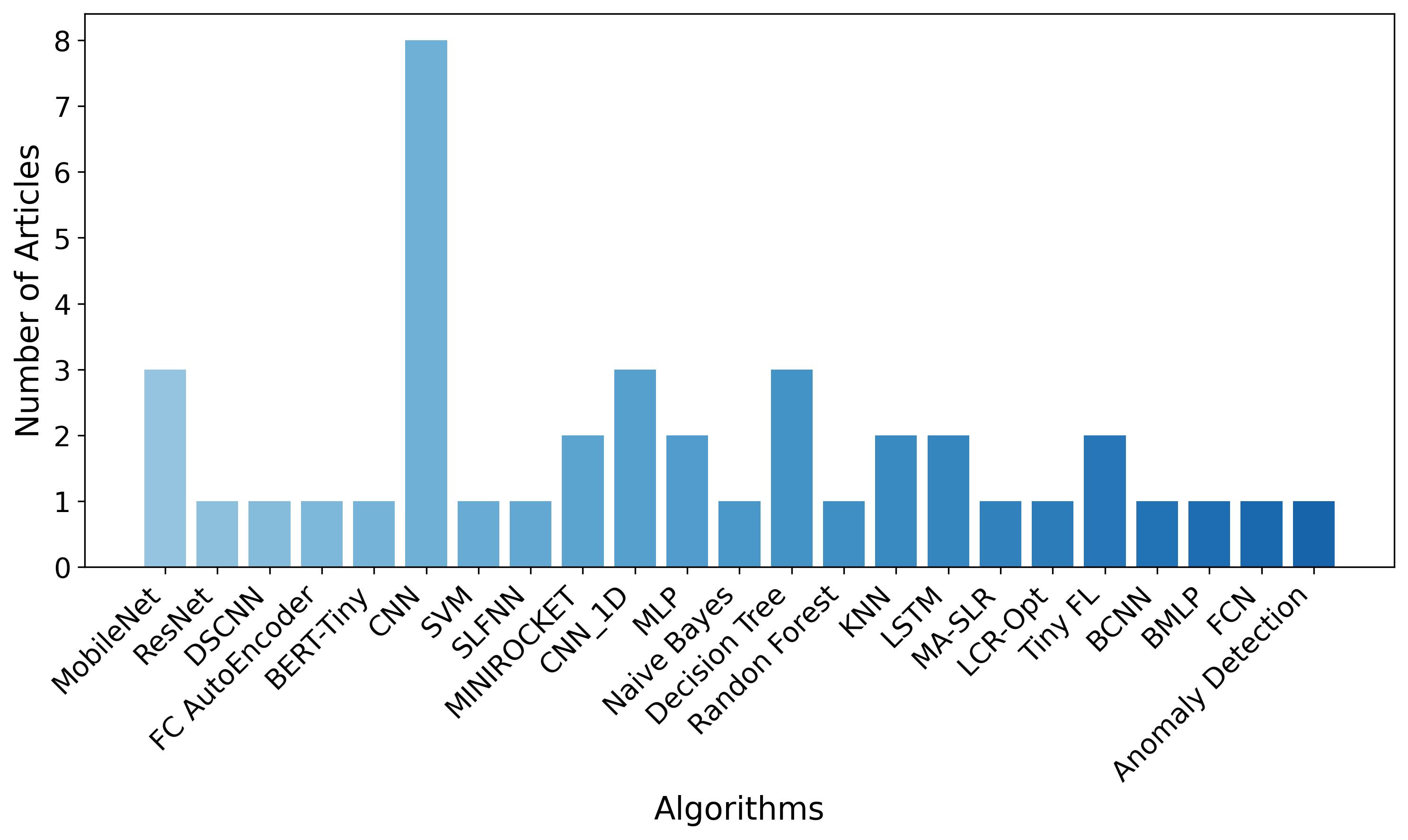}
    \caption{Frequency of \ac{TinyML} Models and Learning Approaches Across Articles. }    \label{fig:frequency_of_TinyML}
\end{figure}

The results from the analysis of the selected articles are shown in Figure~\ref{fig:frequency_of_TinyML}, indicating that the most commonly used \ac{TinyML} algorithm on edge computing devices with limited resources is the \ac{CNN}, appearing in {eight} articles \cite{amin2023embedded, nooruddin2023multi, sabovic2023towards, raza2023glass, arratia2024adaptive, arratia2025towards, lenka2025wisdom,piechocki2022efficient}. Additionally, its variations, such as 1D \ac{CNN} \cite{bian2022exploring, mnif2024combinative, mnif2024ultra}, \ac{BCNN} \cite{raha2025interaxnn} and \ac{DSCNN} \cite{giordano2024tinyforge}, were applied in three, one and one article, respectively. 
Considering \acp{CNN} and their variants as a single family of models, convolutional architectures appeared in {14 out of 25 studies (56.00\%)}. 
It is worth noting that most studies used multiple algorithms, often for comparative purposes, and that a single study may report more than one \ac{TinyML} algorithm, which explains why the frequencies presented in Figure~\ref{fig:frequency_of_TinyML} may exceed the total number of selected studies {(n = 25)}.

Specific techniques for optimizing models for constrained hardware were also identified. For example, MINIROCKET, an efficient method for time-series classification, was used as the main embedded classifier in \cite{giordano2023optimizing} and as part of the asset-activity classification pipeline in \cite{cortesi2024proximity}. Other approaches include \ac{MLP} \cite{wang2020fann, gamazo2023comparison} and its variation \ac{BMLP} \cite{raha2025interaxnn}, MobileNet (a lightweight \ac{CNN} architecture designed for mobile and embedded devices) \cite{giordano2024tinyforge, sabovic2023towards, sanli2025privacy}, \ac{DT} \cite{alwaisi2024securing, alshuhail2025tinyml,lenka2025wisdom}, \ac{LSTM} \cite{hammad2023unsupervised, lenka2025wisdom} and classical machine learning methods such as \ac{KNN} \cite{alwaisi2024securing, al2023tiny}.

Additional algorithms mentioned only once include Residual Networks (ResNet) \cite{giordano2024tinyforge}, \ac{FC} Autoencoder \cite{giordano2024tinyforge}, \ac{BERT}-Tiny \cite{giordano2024tinyforge}, \ac{SVM} \cite{amin2023embedded}, \ac{SLFNN} \cite{amin2023embedded}, \ac{NB} \cite{alwaisi2024securing}, \ac{RF} \cite{alwaisi2024securing}, \ac{MASLR} \cite{al2023tiny}, \ac{LCROpt} \cite{al2023tiny}, \ac{FCN} \cite{lenka2025wisdom}, {and a non-specified TinyML-based anomaly detection approach \cite{alshuhail2025machine}. \ac{TinyFL} appeared in two studies \cite{zheng2024trust, villegas2024optimizing}.}

Various types of \ac{TinyML} algorithms and their variations were used across the studies. The findings reinforce that optimized \acp{CNN} are frequently explored in the literature as viable options for \ac{TinyML} tasks on embedded devices due to their high performance, particularly when optimization techniques such as quantization, pruning and knowledge distillation are applied. These optimized convolutional neural networks are often preferred in energy-constrained environments, as they offer a good balance between performance and resource efficiency when appropriately adapted. 

However, lighter and more specialized models such as MobileNet architectures and the MINIROCKET method have also been adopted as alternatives for resource-constrained deployments. Other approaches, such as \acp{MLP} and classical machine learning algorithms like \ac{KNN}, are also frequently mentioned for their simplicity and robust performance in constrained environments. 
These results indicate that optimized \acp{CNN} are frequently selected when accuracy is a priority, whereas lightweight models such as MINIROCKET, \acp{MLP}, and classical machine learning algorithms may be more suitable for devices with stricter memory and energy constraints.
%These findings suggest that while optimized CNNs are popular for their accuracy and ability to handle complex TinyML tasks, lighter models may be better suited for devices with severe energy and memory constraints. 
It is also worth noting that some promising architectures such as \acp{SNN} were not observed among the selected studies, indicating that these models remain largely underexplored in current \ac{TinyML} energy-focused research.

Table~\ref{tab:rq1_algorithms} summarizes the study-level evidence supporting the frequency-based results presented in Figure~\ref{fig:frequency_of_TinyML}. While the figure highlights how often each algorithm appears, the table shows the application context, algorithm family, and deployment strategy associated with each selected study. This complementary view reveals that algorithm selection is strongly related to the target task and deployment constraints. \ac{CNN}-based models are commonly adopted for image, video, and signal-processing tasks, especially when accuracy is a key requirement. In contrast, lighter approaches such as MINIROCKET, \acp{MLP}, Decision Trees, and statistical models are more frequently associated with scenarios that impose stricter memory, latency, or energy constraints. The table also shows that several studies rely on deployment-oriented strategies, such as quantization, pruning, hardware-aware acceleration, edge processing, and local inference, indicating that \ac{TinyML} algorithm selection is often inseparable from the optimization strategy required by the target device.

\begingroup
\scriptsize
\setlength{\tabcolsep}{3pt}
\renewcommand{\arraystretch}{1.08}

\begin{longtable}{p{1.0cm} p{3.0cm} p{4.0cm} p{4.6cm}}
\caption{Summary of \ac{TinyML} algorithms and deployment characteristics identified in the selected studies.}
\label{tab:rq1_algorithms}\\

\hline
\textbf{Study} & \textbf{Application / Task} & \textbf{Algorithm / Family} & \textbf{Deployment Aspect} \\
\hline
\endfirsthead

\hline
\textbf{Study} & \textbf{Application / Task} & \textbf{Algorithm / Family} & \textbf{Optimization / Deployment Aspect} \\
\hline
\endhead

\hline
\multicolumn{4}{r}{\textit{Continued on next page}}\\
\endfoot

\hline
\endlastfoot

\cite{amin2023embedded} & Robotic hardness classification & \ac{CNN}, \ac{SVM}, \ac{SLFNN} / \ac{CNN}, classical \ac{ML}, neural network & Input-size reduction, memory-aware implementation, and inference optimization \\

\cite{nooruddin2023multi} & Video-based \ac{HAR} & Two-stream multi-resolution \ac{CNN} / \ac{CNN} & \ac{PTQ} and \ac{QAT} for deployment on tiny edge devices \\

\cite{sabovic2023towards} & Battery-less person detection & MobileNet V1 for local and cloud inference / lightweight \ac{CNN} and edge-cloud \ac{ML} & Energy-aware local/remote inference selection \\

\cite{raza2023glass} & Glass fill-level estimation & \ac{CNN} / convolutional neural network & Edge inference to reduce \ac{NBIoT} traffic and sensor energy consumption \\

\cite{arratia2024adaptive} & Ephemeral stream monitoring & \ac{TinyML} vision model / \ac{CNN} image classifier & Adaptive edge retraining to reduce energy and retraining time \\

\cite{arratia2025towards} & Streamflow monitoring & \ac{TinyML} image classifiers / \ac{CNN}-based classifiers & Model selection based on accuracy, memory, and inference cost \\

\cite{lenka2025wisdom} & On-device Wi-Fi sensing & \ac{CNN}, \ac{LSTM}, and \ac{FCN} / neural networks & Context-aware model selection using a \ac{DT}, with pruning, weight clustering, and quantization \\

\cite{piechocki2022efficient} & {Occupancy estimation from thermal images} & {U-Net-based encoder-decoder / \ac{CNN}} & {INT8 quantization and benchmarking on seven \acp{MCU} and edge accelerators} \\

\cite{bian2022exploring} & Gym workout recognition & Residual 1D \ac{CNN} / \ac{CNN} variant & Quantization and parallel execution on GAP8 \\

\cite{mnif2024combinative} & \ac{EIT} hand gesture recognition & 1D \ac{CNN} / \ac{CNN} variant & Combined pruning, quantization, and knowledge distillation \\

\cite{mnif2024ultra} & \ac{EIT} hand gesture classification & Lightweight \ac{CNN} / \ac{CNN} variant & Hardware-aware acceleration and quantization \\

\cite{raha2025interaxnn} & Intermittent \ac{BNN} inference & Binary MLP and binary LeNet / \ac{BNN} variants & Approximate in-memory computing and binary precision \\

\cite{giordano2024tinyforge} & \ac{TinyML} architecture exploration & MobileNet, \ac{DSCNN}, ResNet, \ac{FC} Autoencoder, and \ac{BERT}-Tiny / \ac{CNN}, transformer, and autoencoder architectures & Architecture search and memory hierarchy optimization \\

\cite{giordano2023optimizing} & Asset activity classification & MINIROCKET / time-series classifier & Hyperparameter tuning for memory, accuracy, and energy \\

\cite{cortesi2024proximity} & Asset-operator matching & MINIROCKET-based activity classification, \ac{EKF}-based distance estimation, and cloud matching / time-series classification and signal-processing pipeline & \ac{BLE}-based low-power edge processing \\

\cite{wang2020fann} & \ac{NN} inference on \acp{MCU} & \ac{MLP} / neural network & Parallelization on \ac{RISCV} \ac{PULP} platform \\

\cite{gamazo2023comparison} & Indoor parameter estimation & \ac{MLP} \ac{ANN} / neural network & Distributed edge processing for power reduction \\

\cite{sanli2025privacy} & Privacy-preserving smart city \ac{AI} & MobileNet with federated learning / lightweight neural network and distributed \ac{ML} & Local differential privacy and \ac{FPGA}-based edge processing \\

\cite{alwaisi2024securing} & \ac{IoT} attack detection & \ac{DT}, \ac{KNN}, \ac{RF}, \ac{NB}, \ac{LSTM} / classical \ac{ML} and neural network & Lightweight device-based attack detection \\

\cite{alshuhail2025tinyml} & Structural health monitoring & Pruned \ac{DT} / classical \ac{ML} & Quantization and real-time edge detection on low-power devices \\

\cite{hammad2023unsupervised} & Urban noise anomaly detection & \ac{LSTM} Autoencoder / autoencoder and \ac{LSTM} & Quantization and TensorFlow Lite deployment on ESP32 \\

\cite{al2023tiny} & Data quality validation and imputation & \ac{MASLR}, \ac{KNNEM}, \ac{LCROpt} / statistical and classical \ac{ML} & On-device imputation without cloud dependence \\

\cite{alshuhail2025machine} & Health monitoring and emergency response & Neural networks, statistical anomaly-detection models, and sensor fusion / edge \ac{AI} & Local inference and decentralized emergency response \\

\cite{zheng2024trust} & \ac{TinyFL} trust management in \acp{IUAV} & Tiny wireless \ac{FL} with blockchain trust model / \ac{TinyFL} and distributed \ac{ML} & Trust-aware, low-latency, and energy-efficient \ac{FL} \\

\cite{villegas2024optimizing} & {Federated learning in \ac{IoT} sensor networks} & {Neural network trained with FedAvg and FedProx / \ac{TinyFL}} & {Quantization, pruning, differential privacy, and encrypted model updates} \\

\end{longtable}

\endgroup

\begin{figure*}[h!]
    \centering  \includegraphics[width=1\linewidth]{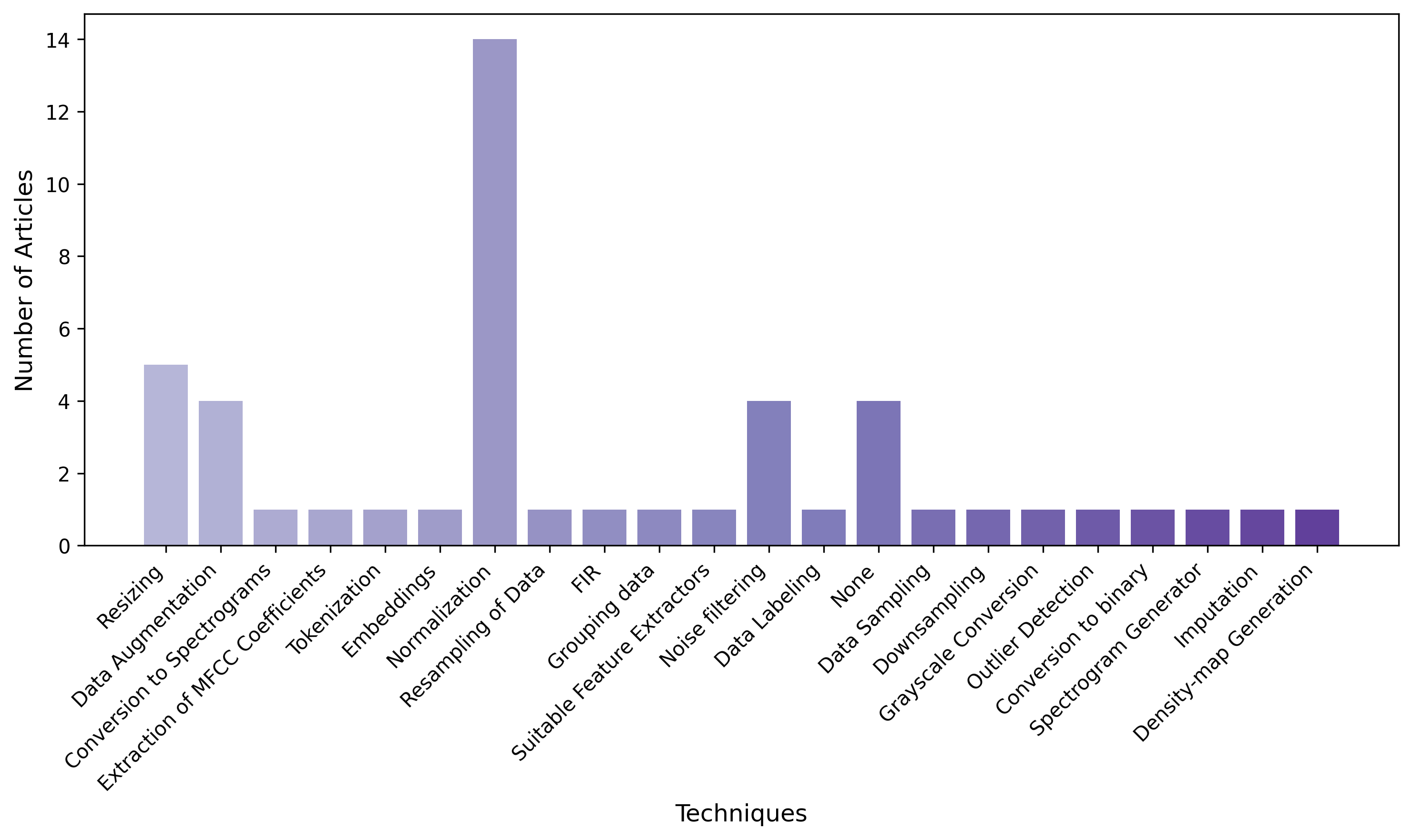}
    \caption{Frequency of Preprocessing Techniques Across Articles. }    \label{fig:frequency_of_Preprocessing_technique}
\end{figure*}

Regarding preprocessing techniques used to prepare data for the chosen algorithm, Figure~\ref{fig:frequency_of_Preprocessing_technique} presents the approaches found in the studies. The most frequently used technique was normalization, {appearing in 14 out of 25} analyzed articles \cite{villegas2024optimizing,giordano2024tinyforge, amin2023embedded, nooruddin2023multi, bian2022exploring, alwaisi2024securing, hammad2023unsupervised, mnif2024ultra, gamazo2023comparison, raha2025interaxnn, giordano2023optimizing, sanli2025privacy, alshuhail2025tinyml, alshuhail2025machine}. Image resizing was used in five applications \cite{giordano2024tinyforge, amin2023embedded, nooruddin2023multi, arratia2024adaptive, arratia2025towards}, while data augmentation \cite{giordano2024tinyforge, wang2020fann, arratia2024adaptive, arratia2025towards} and noise filtering \cite{alwaisi2024securing, nooruddin2023multi, alshuhail2025machine,villegas2024optimizing} were applied in {four articles each.}

The remaining techniques were mentioned only once: spectrogram generator \cite{lenka2025wisdom}, spectrogram conversion \cite{giordano2024tinyforge}, \ac{MFCC} extraction \cite{giordano2024tinyforge}, tokenization \cite{giordano2024tinyforge}, embeddings \cite{giordano2024tinyforge}, data resampling \cite{bian2022exploring}, \ac{FIR} anti-aliasing filtering \cite{bian2022exploring}, data grouping \cite{bian2022exploring}, appropriate feature extraction \cite{wang2020fann}, data labeling \cite{hammad2023unsupervised}, data sampling \cite{al2023tiny}, sampling reduction \cite{sabovic2023towards}, grayscale conversion \cite{sabovic2023towards}, conversion to binary \cite{raha2025interaxnn}, outlier detection \cite{gamazo2023comparison}, {missing-value imputation \cite{villegas2024optimizing}, and density-map generation \cite{piechocki2022efficient}.} Lastly, four articles were classified as ``None'', with three \cite{mnif2024combinative, zheng2024trust, raza2023glass} not reporting preprocessing methods and one \cite{cortesi2024proximity} working directly with raw data.

\subsection{RQ2 - What metrics and methodologies are used to measure and evaluate energy consumption in \ac{TinyML} algorithms on resource-constrained devices?}

To address this research question, we first describe the methodologies used to measure energy consumption, and then discuss the evaluation metrics reported in the selected studies.

\subsubsection{Measurement Methodologies for Energy Consumption}

\begin{figure}[h!]
    \centering    \includegraphics[width=0.8\linewidth]{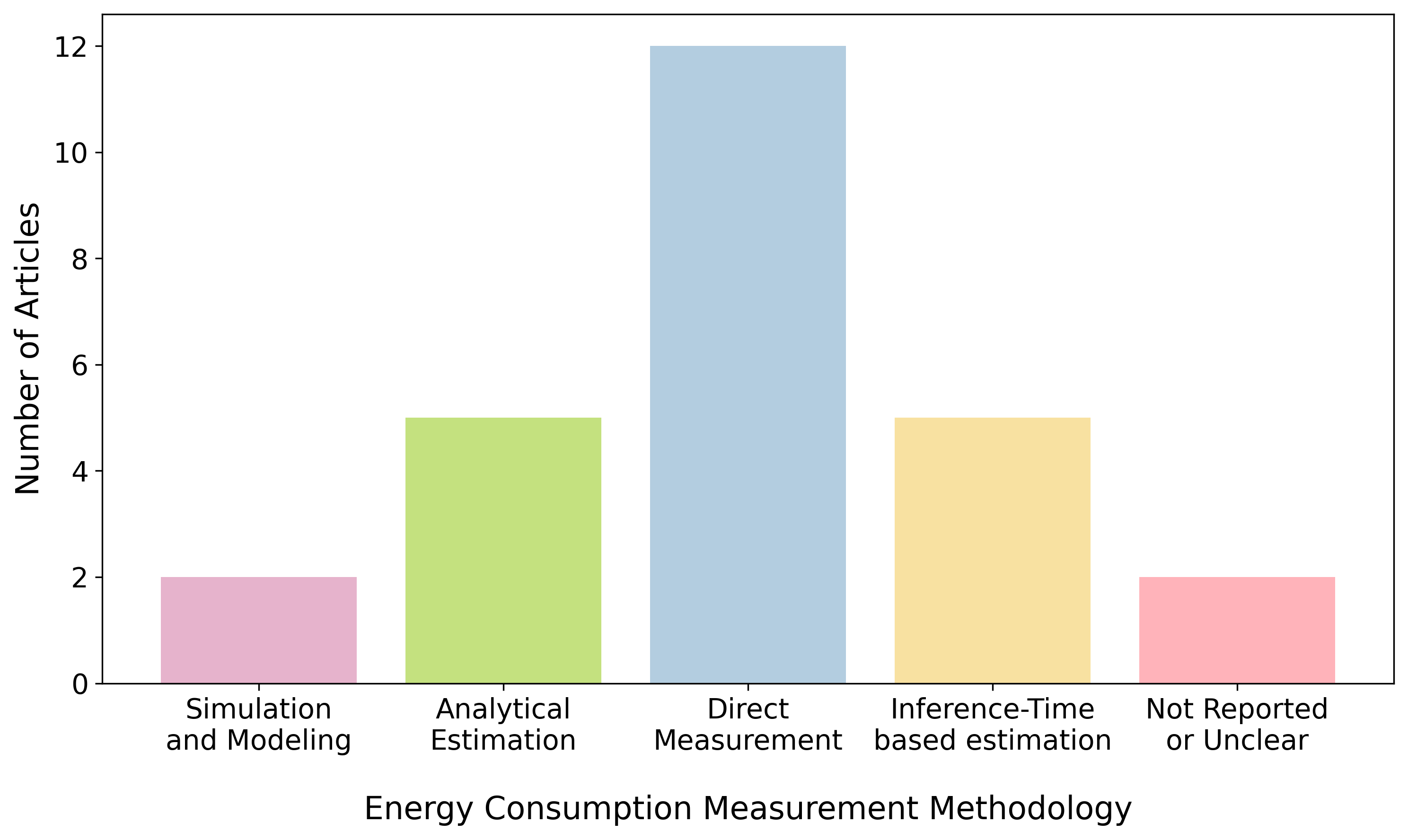}
    \caption{Energy Consumption Measurement Methodology. }    \label{fig:energy_consumption_measurement_methodology}
\end{figure}

Figure~\ref{fig:energy_consumption_measurement_methodology} shows that the methodologies used to measure energy consumption varied across four main measurement approaches and one category for studies in which the methodology was not reported or remained unclear, with some studies combining more than one method. Because these categories are not mutually exclusive, the results should be reported as frequencies rather than percentages. Direct measurement was identified in {twelve} studies, using multimeters, {current sensors,} microcontrollers, Power Profiler Kits, or other instruments to measure current, voltage, or power directly on the device \cite{nooruddin2023multi, alwaisi2024securing, hammad2023unsupervised, al2023tiny, cortesi2024proximity, sabovic2023towards, gamazo2023comparison, mnif2024combinative, arratia2024adaptive, lenka2025wisdom, piechocki2022efficient, villegas2024optimizing}.

For instance, \cite{nooruddin2023multi} proposed a two-stream multi-resolution fusion architecture (context and fovea) for human activity recognition on edge devices with limited resources. In this approach, the context stream receives resized images of the entire scene (e.g., 64×64 pixels), while the fovea stream processes only the central cropped portion (e.g., 32×32 pixels), where subjects typically appear due to camera bias. The extracted features are concatenated and passed to a fully connected layer for classification. The model was quantized using \ac{PTQ} and \ac{QAT}, and then deployed on three edge devices: Arduino Nano 33 \ac{BLE} Sense, Sony Spresense, and Espressif ESP32. Energy consumption was measured using the Nordic \ac{PPKII}. The results showed that while higher resolution images improved metrics such as accuracy and F1-score, they significantly increased energy consumption.

Similarly, \cite{al2023tiny} employed a precision multimeter to measure electrical current (in amperes) consumed by the devices. This current was multiplied by the operating voltage (in volts) to calculate instantaneous power, which was then multiplied by the execution time (in seconds) to obtain total energy consumption in joules. This approach enabled the assessment of energy usage for different algorithms across multiple microcontroller platforms and batch sizes. Among the evaluated algorithms, \ac{MASLR} was the most energy-efficient, outperforming \ac{KNNEM} and \ac{LCROpt}. The tests were conducted on Arduino MKR1000, ESP32, and Raspberry Pi 4B. {In a broader benchmark, \cite{piechocki2022efficient} used a multimeter to measure the current and power of seven \acp{MCU} and edge devices in idle, sleep, and inference states, and combined these measurements with inference time to estimate the energy per work cycle and per day.} These studies serve as representative examples from the set of works selected in the mapping, which employed direct measurement methodologies.

Five studies estimated energy consumption through analytical calculations based on equations \cite{amin2023embedded, zheng2024trust, arratia2025towards, alshuhail2025tinyml, alshuhail2025machine}. In some cases, these estimates relied on values obtained from direct measurements, although this relationship was not always explicitly stated. In the study by \cite{amin2023embedded}, energy consumption on the STM32 Nucleo H745ZI-Q microcontroller was estimated using average current and voltage values obtained through STM32CubeIDE. The STM32 \ac{HAL} provides the function HAL\_GetTick(), which measures elapsed time in milliseconds. Using this function, inference time was determined by calculating the difference between two consecutive readings. Energy consumption was then evaluated using the formula E=P×t, where P is the power consumed and t is inference time. The results showed that \ac{SVM} models had the shortest inference time, resulting in lower energy consumption compared to \ac{SLFNN} and \ac{CNN}, the latter being the most energy-intensive. Zheng et al. \cite{zheng2024trust} proposed a semi-decentralized trust management framework for Tiny \ac{WFL}, enabled by blockchain in \ac{IUAV} systems. The study considered both the computational cost of local training on \ac{UE} and the communication cost to compute total energy consumption as the sum of energy used for local computation and data transmission. The studies by \cite{alshuhail2025tinyml} and \cite{alshuhail2025machine} also used the relation E=P×t and were therefore classified as analytical estimations rather than direct measurements.

Five studies estimated energy consumption using inference or execution time, assessing how long the \ac{AI} model took to process input and generate predictions, then calculating energy usage accordingly \cite{bian2022exploring, wang2020fann, nooruddin2023multi, mnif2024ultra, raza2023glass}. For instance, \cite{bian2022exploring} employed a 1D residual \ac{CNN} to recognize physical exercises on resource-constrained wearable devices. The evaluation was carried out on STM32 Nucleo-144 development boards with Cortex-M4 and Cortex-M7 processors, as well as GAP8, a \ac{RISCV}-based \ac{PULP} platform. The results showed that GAP8, even at maximum clock speed, had the lowest energy consumption per inference. Similarly, Wang et al. \cite{wang2020fann} utilized \acp{MLP} trained with the \ac{FANN} library to execute lightweight, energy-efficient models on \ac{ARM} Cortex-M4 microcontrollers and the ultra-low-power \ac{RISCV}-based \ac{PULP} platform. Their results showed a 69\% reduction in energy consumption on the \ac{RISCV} octa-core platform compared to the Cortex-M4. These studies are examples of this category within the analyzed works.

Simulation and modeling were used in two studies \cite{giordano2024tinyforge, raha2025interaxnn}, while two studies did not clearly specify the measurement methodology used to obtain their energy-related results \cite{giordano2023optimizing, sanli2025privacy}. Giordano et al. \cite{giordano2024tinyforge} proposed a multi-objective optimization strategy to identify Pareto-optimal architectures for energy and area, considering workloads in \ac{TinyML} applications. The authors analyzed Pareto curves comparing an architecture based on \acp{CLA} with conventional approaches. Results indicated that in image classification tasks, the \ac{CLADIMC} architecture could reduce energy consumption by up to 1.6 times while maintaining the same area, or reduce area by up to 3.7 times without increasing energy consumption. In contrast, although \cite{giordano2023optimizing} and \cite{sanli2025privacy} report energy- or power-related results, they do not provide sufficient details about the measurement instrument or procedure to support classification as direct measurement.

Although the reviewed studies report diverse methodologies for evaluating energy consumption, a clear lack of standardization can be observed. Direct measurements using hardware instruments, such as the \ac{PPKII}, precision multimeters, and onboard \ac{MCU} monitors provide the most direct device-level assessment, as they can capture variations in current and voltage during execution. However, they require additional instrumentation, increase experimental complexity, and often depend on specific hardware support. Analytical estimations, which compute energy using equations (e.g., E=P×t), are simpler to implement and do not require external hardware, but their accuracy heavily depends on the reliability of the current, voltage, and timing data obtained from development tools (e.g., STM32CubeIDE or \ac{HAL} functions). Inference-time-based estimates may be less precise when they assume a fixed power consumption profile and ignore dynamic variations that occur during execution. Finally, simulation-based methods allow evaluation at design time, without physical devices, but their results can deviate significantly from real-world deployments due to modeling assumptions.

In addition to methodological differences, the metrics used to report energy consumption also vary considerably. While some studies present total energy in joules (J) or millijoules (mJ) per experiment, others report average power (mW), energy per inference, or aggregated consumption over time. This heterogeneity limits direct comparisons across studies and prevents the establishment of benchmarks. Metrics such as energy per inference and energy efficiency normalized by accuracy are particularly relevant for \ac{TinyML}, as they capture the trade-off between performance and autonomy on resource-constrained devices. These findings highlight the need for greater uniformity in measurement protocols and metrics. Establishing standard evaluation procedures, such as reporting energy per inference, measurement environment details, and the precision of tools, would facilitate reproducibility and improve the comparability of results in future research.

Table~\ref{tab:rq2_measurement_summary} presents the metrics, assumptions, and limitations associated with each measurement category. The results indicate that direct measurement is the most common approach and provides a more realistic assessment of energy consumption. However, comparisons across studies remain challenging because they report different measurements, including current, voltage, power, energy per inference, inference time, memory usage, and modeled energy estimates. This heterogeneity indicates that energy-aware \ac{TinyML} studies still lack a standardized reporting protocol, especially regarding measurement setup, device configuration, and normalized metrics such as energy per inference.

\begin{table*}[!h]
\centering
\caption{Summary of energy measurement methodologies and reported metrics in the selected studies.}
\label{tab:rq2_measurement_summary}
\scriptsize
\setlength{\tabcolsep}{3pt}
\renewcommand{\arraystretch}{1.10}
\begin{tabular}{p{2.0cm} p{3.0cm} p{3.8cm} p{4.2cm}}
\hline
\textbf{Measurement Methodology} & \textbf{Representative Studies} & \textbf{Reported Energy / Resource Metrics} & \textbf{Main Characteristics and Limitations} \\
\hline

Direct measurement & \cite{nooruddin2023multi, alwaisi2024securing, hammad2023unsupervised, al2023tiny, cortesi2024proximity, sabovic2023towards, gamazo2023comparison, mnif2024combinative, arratia2024adaptive, lenka2025wisdom, piechocki2022efficient, villegas2024optimizing} & Current, voltage, power, energy consumption, energy per inference, {energy per work cycle,} memory usage, execution time & Provides the most realistic assessment because it captures device-level consumption during execution. However, it requires measurement instruments, careful experimental setup, and hardware-specific procedures. \\

Analytical estimation & \cite{amin2023embedded, zheng2024trust, arratia2025towards, alshuhail2025tinyml, alshuhail2025machine} & Estimated energy, power-time product, computation cost, communication cost, training or inference time & Easier to apply and useful when direct instrumentation is unavailable. However, accuracy depends on the assumptions, equations, and reliability of current, voltage, and timing estimates. \\

Inference-time-based estimation & \cite{bian2022exploring, wang2020fann, nooruddin2023multi, mnif2024ultra, raza2023glass} & Inference time, latency, energy per inference, power consumption & Useful for comparing deployment efficiency across algorithms and platforms. However, it may oversimplify energy behavior when power varies dynamically during execution. \\

Simulation and modeling & \cite{giordano2024tinyforge, raha2025interaxnn} & Energy per inference, latency, area, forward progress, energy benefits & Enables design-space exploration before physical deployment. However, results depend on modeling assumptions and may differ from real-world measurements. \\

Not reported or unclear & \cite{giordano2023optimizing, sanli2025privacy} & N/R & Limits reproducibility and makes comparison with other \ac{TinyML} deployments difficult. \\

\hline
\end{tabular}
\end{table*}

\subsubsection{Evaluation Metrics}

\begin{figure}[h!]
    \centering  \includegraphics[width=1\linewidth]{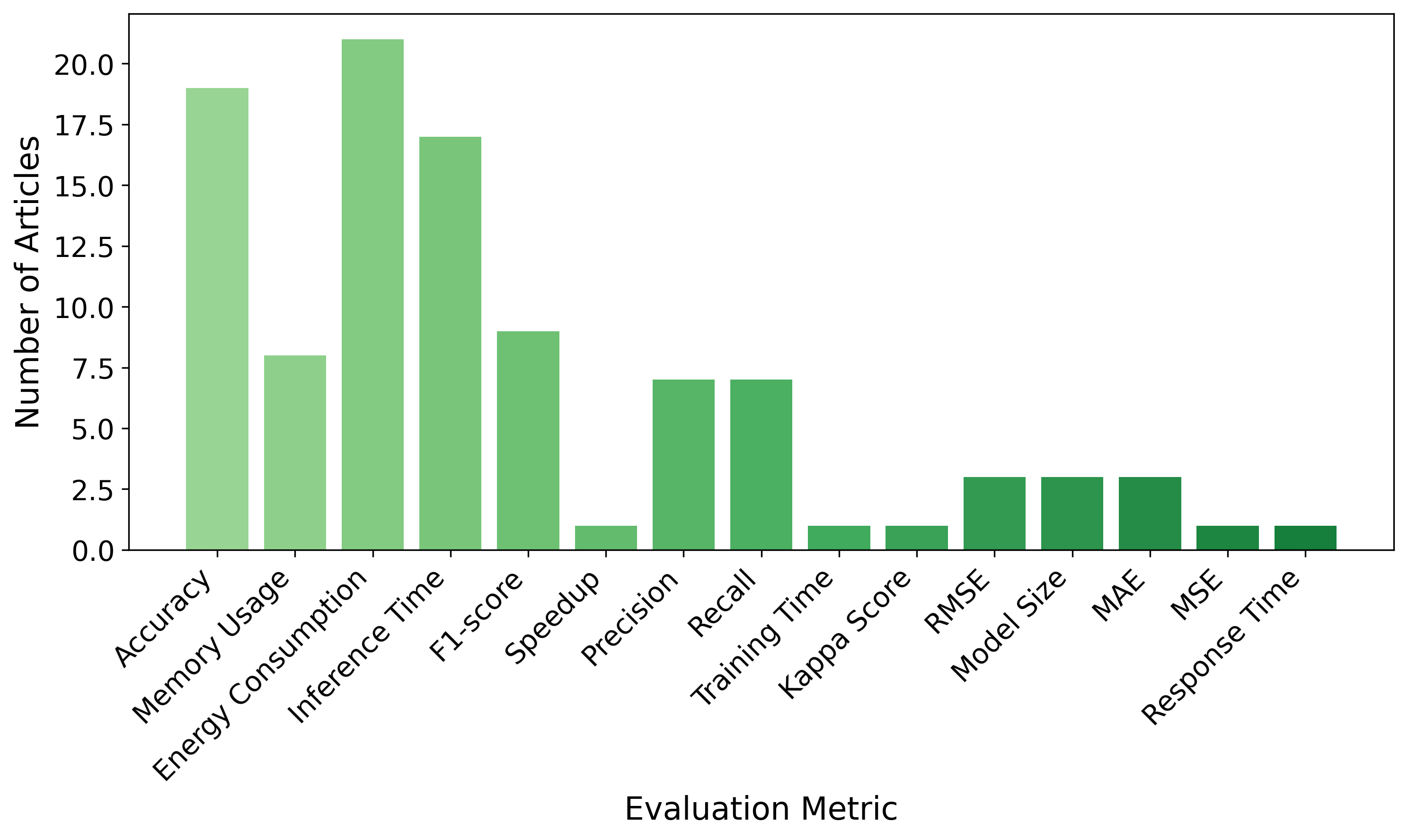}
    \caption{Frequency of Evaluation Metrics in Articles. }    \label{fig:frequency_of_evaluation_metrics_in_articles}
\end{figure}

In addition to measuring energy consumption, the studies also evaluated model performance using various metrics, selected according to the objectives and application context of each study. Figure~\ref{fig:frequency_of_evaluation_metrics_in_articles} shows the frequency of each metric used in the reviewed articles. The analysis reveals that studies prioritize energy consumption and accuracy, along with metrics essential for edge devices, such as memory usage and inference time.

Energy consumption was the most frequently used evaluation metric, appearing in {twenty-one} studies
\cite{amin2023embedded, giordano2023optimizing, bian2022exploring, wang2020fann, alwaisi2024securing, hammad2023unsupervised, mnif2024combinative, al2023tiny, cortesi2024proximity, sabovic2023towards, zheng2024trust, mnif2024ultra, giordano2024tinyforge, raha2025interaxnn, arratia2024adaptive, sanli2025privacy, alshuhail2025tinyml, lenka2025wisdom, alshuhail2025machine, piechocki2022efficient, villegas2024optimizing}.
Accuracy followed closely, being considered in {nineteen} studies
\cite{giordano2024tinyforge, amin2023embedded, giordano2023optimizing, bian2022exploring, wang2020fann, alwaisi2024securing, hammad2023unsupervised, nooruddin2023multi, raza2023glass, gamazo2023comparison, raha2025interaxnn, arratia2024adaptive, sanli2025privacy, alshuhail2025tinyml, arratia2025towards, lenka2025wisdom, alshuhail2025machine, piechocki2022efficient, villegas2024optimizing},
as it indicates the proportion of correct predictions made by the model. Inference time was also adopted in {seventeen} papers
\cite{giordano2024tinyforge, amin2023embedded, giordano2023optimizing,
bian2022exploring, wang2020fann, hammad2023unsupervised,
nooruddin2023multi, mnif2024combinative, al2023tiny,
sabovic2023towards, mnif2024ultra, raza2023glass,
sanli2025privacy, alshuhail2025tinyml, lenka2025wisdom,
alshuhail2025machine, piechocki2022efficient}, being particularly relevant for real-time applications, as it measures how quickly a model produces an output from a given input.

The F1-score appeared in {nine} papers \cite{hammad2023unsupervised, alwaisi2024securing, mnif2024ultra, nooruddin2023multi, cortesi2024proximity, gamazo2023comparison, arratia2024adaptive, arratia2025towards, piechocki2022efficient}, as it provides a harmonic mean between precision and recall. Memory usage was used in eight studies \cite{giordano2023optimizing, amin2023embedded, alwaisi2024securing, hammad2023unsupervised, nooruddin2023multi, mnif2024ultra, gamazo2023comparison, lenka2025wisdom}, assessing the amount of memory required during model execution, which is important for resource-constrained environments. Precision \cite{alwaisi2024securing, hammad2023unsupervised, mnif2024combinative, nooruddin2023multi, cortesi2024proximity, arratia2025towards,alshuhail2025machine} and recall \cite{alwaisi2024securing, hammad2023unsupervised, nooruddin2023multi, cortesi2024proximity, mnif2024ultra, arratia2025towards, alshuhail2025machine} were reported in seven studies each, reflecting a model’s ability to correctly identify relevant instances. Other metrics, such as \ac{RMSE} \cite{hammad2023unsupervised, al2023tiny, gamazo2023comparison}, training time \cite{alwaisi2024securing}, \ac{MAE} \cite{al2023tiny, gamazo2023comparison, piechocki2022efficient}, {mean squared error \cite{piechocki2022efficient}, response time \cite{villegas2024optimizing},} speedup \cite{wang2020fann}, Kappa Score \cite{hammad2023unsupervised}, forward  progress \cite{raha2025interaxnn} and model size \cite{mnif2024combinative, piechocki2022efficient, villegas2024optimizing}, appeared in four or fewer studies. These findings highlight that metric selection varies depending on the objectives of each study and the limitations of the target deployment environment. Although this mapping primarily focuses on inference and deployment on resource-constrained devices, training time was reported as a complementary metric alongside inference- and resource-related measurements, rather than as the central focus of the evaluated studies.

Table~\ref{tab:rq2_evaluation_metrics} groups the reported metrics according to their role in \ac{TinyML} evaluation. This organization shows that energy-aware \ac{TinyML} studies do not rely only on predictive metrics, such as accuracy and F1-score, but also on deployment-oriented metrics, including energy consumption, inference time, memory usage, and model size. Therefore, the evaluation of \ac{TinyML} algorithms requires a multi-dimensional perspective, since a model with high accuracy may still be unsuitable for deployment if it exceeds the energy, latency, or memory constraints of the target device.

\begin{table*}[!ht]
\centering
\caption{Summary of evaluation metrics used in energy-aware \ac{TinyML} studies.}
\label{tab:rq2_evaluation_metrics}
\scriptsize
\setlength{\tabcolsep}{3pt}
\renewcommand{\arraystretch}{1.10}
\begin{tabular}{p{2.0cm} p{3.0cm} p{4.0cm} p{4.0cm}}
\hline
\textbf{Metric Group} & \textbf{Metrics} & \textbf{Purpose in \ac{TinyML} Evaluation} & \textbf{Interpretation for Resource-Constrained Devices} \\
\hline

Energy and power metrics & Energy consumption, power consumption, energy per inference & Measure the energy cost of executing a model or system task & Essential for estimating battery lifetime, feasibility of deployment, and suitability for energy-harvesting or low-power devices \\

Timing metrics & Inference time, latency, execution time, training time{, response time} & Measure how long the model takes to process data or complete training/inference & Important for real-time applications, where delayed inference can compromise responsiveness or safety \\

Predictive performance metrics & Accuracy, F1-score, precision, recall, Kappa score & Measure the predictive quality of the model & Necessary to evaluate whether energy savings are achieved without unacceptable degradation in model performance \\

Memory and model footprint metrics & Memory usage, \ac{RAM}, flash, model size & Measure the storage and runtime memory requirements of the model & Critical for deployment on \acp{MCU} and edge devices with limited \ac{RAM}, flash, and computational capacity \\

Regression and error metrics & \ac{MAE}, {mean squared error,} \ac{RMSE} & Measure numerical prediction or estimation errors & Relevant for sensing, imputation, {density estimation,} and environmental estimation tasks where outputs are continuous values rather than classes \\

System-level efficiency metrics & Speedup, forward progress, communication cost & Measure broader execution efficiency beyond model accuracy & Useful for comparing hardware acceleration, distributed execution, intermittent computing, and edge-cloud trade-offs \\

\hline
\end{tabular}
\end{table*}

\subsection{RQ3 - What Technical Factors Influence Energy Consumption Measurements in \ac{TinyML} Algorithms Deployed on Edge Devices?}

\begin{figure}[h!]
    \centering    \includegraphics[width=1\linewidth]{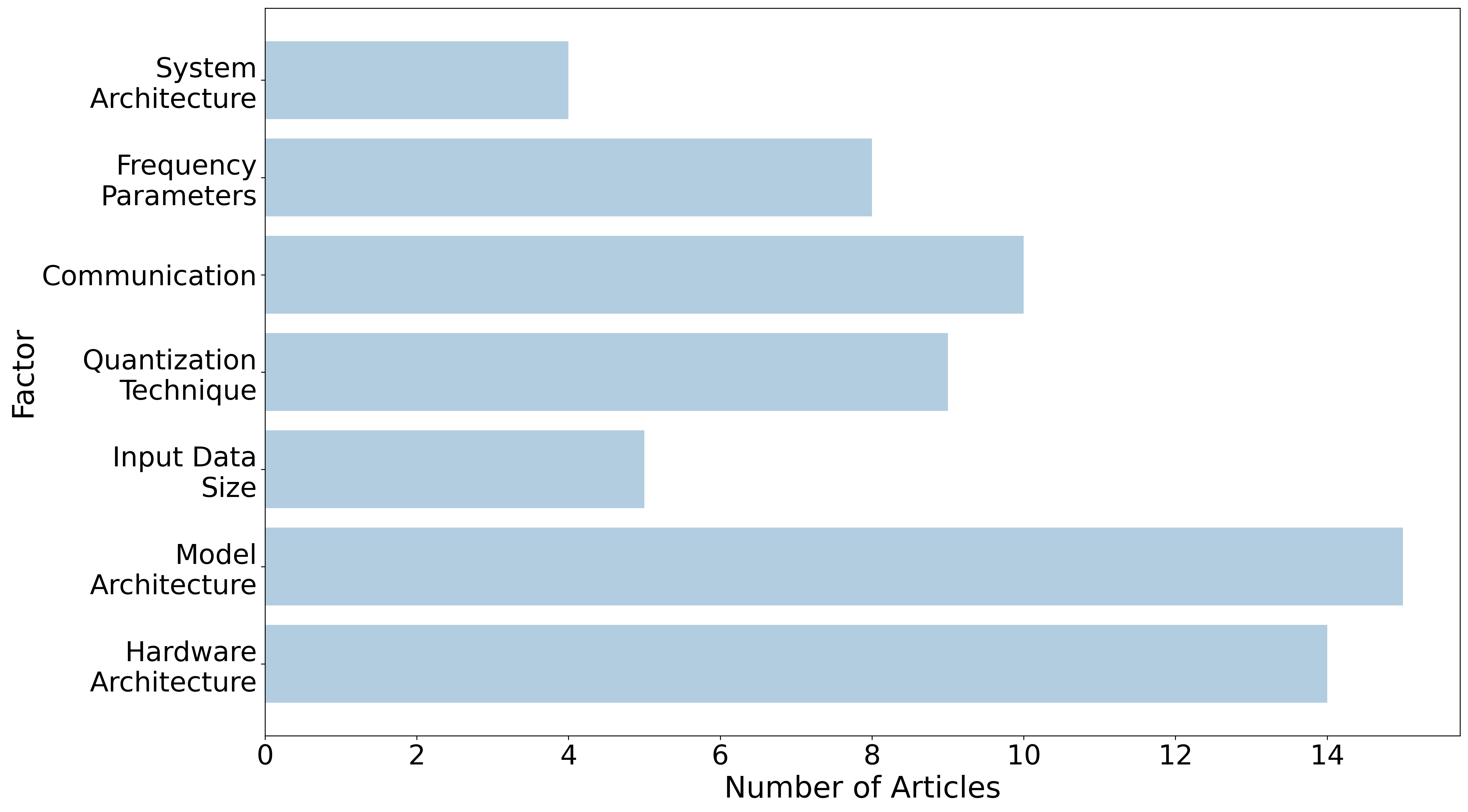}
    \caption{Factors Influencing Energy Consumption in \ac{TinyML}. }  \label{fig:factors_influencing_energy_consumption_in_tinyML}
\end{figure}

Several factors can influence the energy consumption of \ac{TinyML} algorithms, and understanding them is essential for identifying effective optimization strategies. These factors may range from hardware characteristics to methodological choices and model architecture design. Figure~\ref{fig:factors_influencing_energy_consumption_in_tinyML} presents the factors identified in the {25} articles analyzed in this systematic mapping study. Because a single study may report more than one factor, the categories are not mutually exclusive, and their frequencies may exceed the total number of selected studies.

The data highlight the importance of selecting the appropriate hardware architecture and model architecture for devices with limited energy resources. Model architecture \cite{giordano2024tinyforge, amin2023embedded, wang2020fann, alwaisi2024securing, mnif2024combinative, al2023tiny, nooruddin2023multi, raza2023glass, gamazo2023comparison, raha2025interaxnn, alshuhail2025tinyml, arratia2025towards, lenka2025wisdom, alshuhail2025machine, arratia2024adaptive} was identified as an influencing factor in 15 articles, while hardware architecture was addressed in {14} articles \cite{giordano2024tinyforge, bian2022exploring, wang2020fann, alwaisi2024securing, mnif2024combinative, al2023tiny, nooruddin2023multi, mnif2024ultra, gamazo2023comparison, raha2025interaxnn, alshuhail2025tinyml, lenka2025wisdom, alshuhail2025machine, piechocki2022efficient}. For hardware architecture, the most cited factor was the type of processor used. Regarding model architecture, energy consumption is influenced by aspects such as model size and type, architecture design, number of operations, parameter count, number of inference runs required to obtain a result, and inference time.

Communication method and frequency and timing parameters were cited in {ten and eight} articles \cite{cortesi2024proximity, sabovic2023towards, zheng2024trust, gamazo2023comparison, sanli2025privacy, alshuhail2025tinyml, arratia2025towards, alshuhail2025machine, villegas2024optimizing, raza2023glass} and \cite{giordano2024tinyforge, giordano2023optimizing, bian2022exploring, zheng2024trust, raha2025interaxnn, alshuhail2025tinyml, arratia2025towards, piechocki2022efficient}, respectively. Frequency and timing parameters include processor clock rate, sampling frequency, {\ac{CPU} timing, and duty-cycle duration}. In addition, quantization techniques \cite{nooruddin2023multi, mnif2024ultra, raza2023glass, raha2025interaxnn, alshuhail2025tinyml, lenka2025wisdom, piechocki2022efficient, villegas2024optimizing, hammad2023unsupervised} and input data size \cite{amin2023embedded, nooruddin2023multi, gamazo2023comparison, arratia2025towards, arratia2024adaptive} were mentioned in {nine} and five studies, respectively. System architecture aspects, such as memory hierarchy, dataflow organization, and task scheduling, were reported as influencing factors in four articles \cite{giordano2024tinyforge, raha2025interaxnn, sabovic2023towards, arratia2025towards}.

In addition to listing the factors, the selected studies also provide evidence of how some of them affect energy consumption in practice. In particular, the architecture of the \ac{MCU} was shown to have a substantial impact in the evaluated configurations. Devices based on parallel or specialized architectures, such as GAP8 \cite{bian2022exploring} and \ac{CLADIMC} \cite{giordano2024tinyforge}, reported lower energy consumption in their respective comparisons with general-purpose processing architectures. 
%These gains were associated with support for reduced-precision operations, parallel execution, and specialized memory hierarchies that reduce data movement. 
These gains were associated with {parallel execution and specialized memory hierarchies} that reduce data movement.
For example, GAP8 achieved lower energy per inference than the evaluated Cortex-M4 and Cortex-M7 platforms for the 1D \ac{CNN} workload in \cite{bian2022exploring}. Similarly, the modeled \ac{CLADIMC} architecture provided favorable energy--area trade-offs for the workloads evaluated in \cite{giordano2024tinyforge}. {On the other hand, processing speed alone did not determine energy efficiency. In a benchmark of the same quantized model on seven devices, the ESP32-based board consumed about 22 times less energy per work cycle than the fastest \ac{MCU} evaluated, mainly because of its lower power draw in sleep and inference states~\cite{piechocki2022efficient}.} These findings suggest that hardware--algorithm co-design is essential for achieving energy-efficient \ac{TinyML} inference and should be considered when comparing consumption across devices.

Table~\ref{tab:rq3_technical_factors} organizes the identified factors according to their practical effect on \ac{TinyML} energy consumption. This view shows that energy behavior is not determined only by the selected algorithm, but by the interaction among hardware architecture, model complexity, communication strategy, input characteristics, and system-level execution decisions. In particular, the table reinforces that hardware-aware design and model optimization are closely related: the same algorithm may present different energy profiles depending on processor type, memory hierarchy, numerical precision, and data movement patterns. Therefore, energy-aware \ac{TinyML} deployment requires a co-design perspective that considers both algorithmic and platform-level constraints.

\begin{table*}[!ht]
\centering
\caption{Summary of technical factors influencing energy consumption in \ac{TinyML} deployments.}
\label{tab:rq3_technical_factors}
\scriptsize
\setlength{\tabcolsep}{3pt}
\renewcommand{\arraystretch}{1.10}
\begin{tabular}{p{2.5cm} p{4.0cm} p{4.0cm} p{2.5cm}}
\hline
\textbf{Technical Factor} & \textbf{Examples Identified in the Studies} & \textbf{Energy-Related Impact} & \textbf{Representative Studies} \\
\hline

Hardware architecture & \ac{MCU} type, processor architecture, accelerators, memory hierarchy, heterogeneous or parallel platforms & Determines how efficiently operations are executed and how much energy is spent on computation and memory access & \cite{giordano2024tinyforge,bian2022exploring,wang2020fann,raha2025interaxnn,lenka2025wisdom,piechocki2022efficient} \\

Model architecture & Model type, number of parameters, number of operations, model size, inference runs, layer organization & More complex models can increase computation, memory access, latency, and energy per inference & \cite{amin2023embedded, nooruddin2023multi, mnif2024combinative, al2023tiny, arratia2025towards, raha2025interaxnn} \\

Communication method & \ac{BLE}, \ac{NBIoT}, Wi-Fi, {LoRa,} edge-cloud transmission, federated updates, \ac{UAV} communication & Communication can dominate energy consumption, especially when raw data or model updates are transmitted frequently & \cite{cortesi2024proximity,sabovic2023towards,zheng2024trust,sanli2025privacy,raza2023glass,villegas2024optimizing} \\

Frequency and timing parameters & Processor clock, sampling frequency, \ac{CPU} timing, inference latency, task execution time{, duty-cycle duration} & Higher frequency may reduce latency but can increase instantaneous power consumption and total energy depending on workload duration & \cite{bian2022exploring, giordano2024tinyforge, raha2025interaxnn, alshuhail2025tinyml, arratia2025towards, piechocki2022efficient} \\

Quantization and precision & 8-bit arithmetic, binary precision, quantized neural networks, \ac{PTQ}, \ac{QAT}, and quantized TensorFlow Lite models & Reduces memory footprint, arithmetic cost, and data movement, often lowering inference energy with limited accuracy loss & \cite{nooruddin2023multi, mnif2024ultra, raha2025interaxnn, lenka2025wisdom, hammad2023unsupervised, villegas2024optimizing} \\

Input data characteristics & Image resolution, input size, sensor sampling, window size, data dimensionality & Larger inputs increase memory access, processing time, and number of operations, while reduced inputs can lower energy consumption & \cite{amin2023embedded,nooruddin2023multi,gamazo2023comparison,arratia2025towards,arratia2024adaptive} \\

System-level architecture & Task scheduling, dataflow organization, edge-cloud partitioning, memory placement, local processing & Can reduce unnecessary computation, avoid redundant data movement, and improve energy-aware execution across the system stack & \cite{sabovic2023towards, giordano2024tinyforge, raha2025interaxnn, arratia2025towards} \\

\hline
\end{tabular}
\end{table*}

\subsection{RQ4 - What Techniques Have Been Proposed to Optimize Energy Consumption in \ac{TinyML} Algorithms on Edge Devices Based on Measurements?}

Before discussing optimization strategies, it is important to understand how different \ac{TinyML} algorithms behave in terms of energy efficiency. This context clarifies why specific techniques are necessary and how they address the limitations of each algorithm.

\subsubsection{\ac{TinyML} Algorithms and Their Energy Efficiency}
\label{sec:algorithms_energy}

Among the \ac{TinyML} algorithms identified, \acp{CNN} and their variants (1D \ac{CNN}, 
\ac{BCNN} and \ac{DSCNN}), MINIROCKET, \ac{KNN}, and \ac{MLP} were the most frequently adopted. 
These models support diverse tasks (e.g.: image, video, and time-series 
classification) and run on a wide range of hardware, from low-power 
microcontrollers to Systems on a Chip (SoCs) with dedicated accelerators.

For \acp{CNN}, energy per inference varied substantially across platforms and 
configurations. On general-purpose \acp{MCU}, values ranged from 
331~$\mu$J/inf to 657~$\mu$J/inf on the STM32 H745ZI-Q depending on 
input size, with the \ac{CNN} achieving the best accuracy ($>$98\%) among the 
tested algorithms~\cite{amin2023embedded}. However, this accuracy came 
at a cost: the \ac{CNN} consumed approximately 52$\times$ more energy than 
an \ac{SVM} solving the same task on the same device (657~$\mu$J/inf vs.\ 12.51~$\mu$J/inf). 
Additionally, enabling the on-chip L1-cache reduced \ac{CNN} energy consumption 
by up to 80\%, from 657~$\mu$J to 132~$\mu$J/inf, without any 
modification to the model~\cite{amin2023embedded}. Similarly, reducing 
the input size by half (from N=80 to N=40 samples) cut energy 
consumption by approximately 50\% with only a 0.4\% accuracy 
drop~\cite{amin2023embedded}. 
%On SoCs such as GAP8, optimized with parallelism and quantization,  consumption dropped to 0.41~mJ/inf, approximately 12--20$\times$ more efficient than the Cortex-M4 and Cortex-M7 platforms evaluated in the same study~\cite{bian2022exploring}.
On SoCs such as GAP8, {which execute inference in parallel on an eight-core RISC-V cluster,} consumption dropped to 0.41~mJ/inf, approximately 12--20$\times$ {less} than the Cortex-M4 and Cortex-M7 platforms evaluated in the same study~\cite{bian2022exploring}. {Because the same 8-bit quantized model was deployed on all three platforms, this gain is attributable to the hardware architecture and deployment toolchain rather than to quantization.}

Deploying the same quantized \ac{CNN} model across three commodity devices 
revealed that higher clock speed reduces inference latency but increases 
power draw: the ESP32 (240~MHz) achieved the lowest inference time 
($\sim$400~ms) but the highest power consumption ($\sim$190~mW), 
the Sony Spresense (156~MHz) presented intermediate values 
($\sim$1100~ms, $\sim$115~mW), while the Arduino Nano 33 \ac{BLE} (64~MHz) 
showed the lowest power consumption ($\sim$50~mW) at the cost of 
higher latency ($\sim$2000~ms)~\cite{nooruddin2023multi}. Furthermore, reducing 
input resolution from a (64,32) to a (32,16) two-stream configuration 
reduced power consumption by approximately 58\% and model size by 66\%, 
while accuracy on the KTH dataset dropped by less than 
0.1\%~\cite{nooruddin2023multi}. {A similar trade-off was observed for a quantized U-Net-based model for people counting. A Raspberry Pi 4B with a Coral USB accelerator achieved the lowest inference time ($\sim$0.6~ms) but required 93.5~J per 30-second work cycle, whereas low-power \acp{MCU} such as the Arduino Nano 33 \ac{BLE} Sense required only 2.45~J per cycle despite an inference time of about 1.4~s~\cite{piechocki2022efficient}.}

\acp{MLP} also delivered competitive results, achieving up to 69\% reduction 
in energy consumption on parallel \ac{RISCV} architectures compared to a 
single-core Cortex-M4~\cite{wang2020fann}. In the data-quality validation and imputation study, the \ac{KNN}-based approach consumed more energy than \ac{MASLR}, although its cost could be reduced by limiting the number of samples used during execution~\cite{al2023tiny}. MINIROCKET, optimized 
for time-series classification, demonstrated ultra-low average system 
power below 15~$\mu$W with a memory footprint of only 7~kB flash and 
3~kB \ac{RAM}~\cite{giordano2023optimizing}.

Across the studies that explicitly evaluated optimization techniques, quantization, pruning, knowledge distillation, input-size reduction, caching, and hardware-aware execution produced substantial energy or resource savings. Reported results included reductions of up to 80\% from on-chip caching~\cite{amin2023embedded}, 69\% through parallel \ac{RISCV} execution~\cite{wang2020fann}, {52\% through combined model compression~\cite{mnif2024combinative}, and 25--30\% in power consumption through quantization in a federated setting~\cite{villegas2024optimizing}}. However, heterogeneity in methodologies, hardware, and application domains prevents direct comparisons.

\subsubsection{Optimization Techniques}

The optimization techniques identified in the literature can be systematically classified into four primary categories, as summarized in Table~\ref{tab:optimization_taxonomy}. This taxonomy provides a structured framework for selecting appropriate strategies according to system constraints and objectives.  Because a study may apply more than one optimization technique, the categories and subcategories are not mutually exclusive. 

\begin{table*}[!t]
\centering
\caption{Taxonomy of Energy Optimization Techniques.}
\label{tab:optimization_taxonomy}

\begingroup
\scriptsize
\setlength{\tabcolsep}{3pt}
\renewcommand{\arraystretch}{1.15}

\begin{tabular}{@{}p{2.5cm}p{2.8cm}p{5.2cm}p{4.0cm}@{}}
\hline

\textbf{Category} &
\textbf{Subcategory} &
\textbf{Key Techniques} &
\textbf{Representative Studies} \\
\hline

\textbf{Model Compression}
& Quantization
& PTQ, QAT, integer quantization, and binary quantization
& \cite{nooruddin2023multi,bian2022exploring,mnif2024combinative,
mnif2024ultra,raha2025interaxnn,alshuhail2025tinyml,
hammad2023unsupervised,lenka2025wisdom,piechocki2022efficient,villegas2024optimizing} \\

& Pruning
& Structured and unstructured pruning
& \cite{mnif2024combinative,alshuhail2025tinyml,
lenka2025wisdom,villegas2024optimizing} \\

& Knowledge Distillation
& Teacher--student frameworks
& \cite{mnif2024combinative} \\

& Weight Clustering
& Weight sharing and pruning--clustering combinations
& \cite{lenka2025wisdom} \\
\hline

\textbf{Hardware-Aware Design}
& Architecture and Memory Design
& Memory configurations, compute-in-memory, and Pareto-optimal design
& \cite{giordano2024tinyforge,raha2025interaxnn} \\

& Parallel and Hardware-Aware Execution
& Parallel execution, specialized processors, and hardware acceleration
& \cite{bian2022exploring,mnif2024ultra,wang2020fann} \\
\hline

\textbf{System \& Communication}
& Energy-Aware Local--Cloud and Adaptive Execution
& Local/remote inference selection and adaptive retraining
& \cite{sabovic2023towards,arratia2024adaptive} \\

& Protocol Optimization
& BLE-based communication and scheduled transmission
& \cite{cortesi2024proximity} \\

& Federated and Distributed Resource Allocation
& Computation, communication, aggregation, and trust-aware allocation
& \cite{zheng2024trust,villegas2024optimizing} \\

& {Duty Cycling and Low-Power Modes}
& {Periodic inference with sleep modes between work cycles}
& \cite{piechocki2022efficient} \\
\hline

\textbf{Data \& Processing}
& Input Optimization
& Input-size reduction, resolution reduction, sampling reduction, and caching
& \cite{amin2023embedded,nooruddin2023multi,
sabovic2023towards} \\

& Model and Hyperparameter Selection
& Selection based on energy, memory, accuracy, and inference cost
& \cite{arratia2025towards,giordano2023optimizing} \\

& Dataflow and Memory Optimization
& On-chip cache/SRAM, CLADIMC, compute-in-memory, and reduced data movement
& \cite{amin2023embedded,giordano2024tinyforge,
raha2025interaxnn} \\
\hline

\end{tabular}
\endgroup
\end{table*}

The first category, \textbf{Model Compression}, reduces computational and memory requirements through quantization, pruning, knowledge distillation, and weight clustering. 
% Quantization, often reducing parameter precision from 32-bit to 8-bit, showed clear benefits: for example, an 8-bit quantized residual neural network maintained accuracy with only a 2.3\% drop while reducing inference energy to 0.41~mJ on GAP8, compared to over 5~mJ on Cortex-M4~\cite{bian2022exploring}. 
Quantization, often reducing parameter precision from 32-bit to 8-bit, {enabled deployment on constrained platforms with limited accuracy loss. For example, 8-bit quantization of a residual neural network reduced accuracy by only 1.1 to 2.3 percentage points across three embedded platforms~\cite{bian2022exploring}.}
Pruning removed redundant parameters and, when combined with quantization, reduced model size by 98.8\% (from 871.58 kB to 10.42 kB) with an accuracy trade-off of 8.56\%~\cite{mnif2024combinative}. Knowledge distillation transferred knowledge from larger teacher models to compact student models, enabling lightweight deployment under tight energy budgets~\cite{mnif2024combinative}. Weight clustering was also investigated in WISDOM, both individually and in combination with pruning and quantization, to reduce flash, runtime-memory, and energy requirements~\cite{lenka2025wisdom}. {In a TinyFL setting, compression and quantization reduced model size by about 75\% and power consumption by 25--30\%, with an average accuracy loss of about 1.5\%~\cite{villegas2024optimizing}.} These approaches form a complementary set that balances efficiency with accuracy in constrained devices.

The second category, \textbf{Hardware-Aware Design}, improves efficiency by aligning algorithms with hardware capabilities. The selected studies include architecture and memory design, such as the exploration of \ac{SRAM}, \ac{SLA}, and \ac{CLA} memory configurations and \ac{STDC} and \ac{DIMC} computing approaches in TinyForge~\cite{giordano2024tinyforge}. InterAxNN combines compute-in-memory with approximate binary operations and runtime reconfiguration to increase forward progress under intermittent power~\cite{raha2025interaxnn}. 
% Other studies exploit parallel or specialized architectures, including GAP8 and \ac{RISCV}-based \ac{PULP} platforms, to reduce inference energy~\cite{bian2022exploring,wang2020fann}. These strategies exemplify hardware--algorithm co-design adapted for constrained environments.
Other studies exploit parallel or specialized architectures, including GAP8 and \ac{RISCV}-based \ac{PULP} platforms, to reduce inference energy~\cite{bian2022exploring,wang2020fann}{, with gains of approximately one order of magnitude over single-core Cortex-M platforms (Section~\ref{sec:algorithms_energy})}.
%TinyForge explores memory and compute configurations, while InterAxNN attributes most of its forward-progress gains to compute-in-memory parallelism and reduced memory traffic.

The third category, \textbf{System \& Communication Optimizations}, addresses resource allocation and transmission overhead. Sabovic et al.~\cite{sabovic2023towards} proposed an energy-aware decision mechanism that selects between local and cloud inference according to available energy, accuracy, and execution-time constraints. In the evaluated natural-light scenario, the system favored remote inference under high-energy conditions and local inference when harvested energy was lower. \ac{BLE}-based communication also supported low-power asset and operator monitoring: the SMARTTAG consumed only 15~$\mu$W, enabling an estimated lifetime of more than three years on a 675~mWh battery~\cite{cortesi2024proximity}. In the ECOTRACK wearable, most energy was spent in \ac{BLE} scanning (0.31~Wh during an 8-hour workday), while algorithmic processing represented only 8.2~$\mu$Wh per day~\cite{cortesi2024proximity}. For Tiny Federated Learning, Zheng et al.~\cite{zheng2024trust} formulated a resource-allocation problem that considers local computation, wireless transmission, block validation, learning time, and energy cost. {Villegas-Ch et al.~\cite{villegas2024optimizing} reported that federated training on Cortex-M4 nodes consumed about 33\% less power than a centralized approach (100~mW vs.\ 150~mW), since only model updates were transmitted. Duty cycling was also explored. Piechocki et al.~\cite{piechocki2022efficient} evaluated periodic operation with 30-second and 5-minute work cycles, keeping the devices in a low-power mode between inferences, and reported a daily energy consumption of 2.64~kJ and 2.48~kJ, respectively, on the most efficient board.} These approaches exemplify system- and communication-level optimization aimed at balancing local computation, wireless transmission, latency, and energy consumption in constrained TinyML environments.

The fourth category, \textbf{Data \& Processing Optimizations}, minimizes data movement and access costs. Techniques included resolution reduction (e.g., reducing the two-stream input from (64,32) to (32,16) cut power consumption by approximately 58\%~\cite{nooruddin2023multi}), memory caching (STM32 on-chip L1 cache reduced inference energy by up to 80\%~\cite{amin2023embedded}), and custom dataflows like \ac{CLADIMC}, which stores weights directly on-chip to reduce energy by up to 1.6$\times$ at equal silicon area, or reduce silicon area by up to 3.7$\times$ at equal energy~\cite{giordano2024tinyforge}. These results show that reducing data transfers is often as critical as reducing computation.

This taxonomy reveals four complementary domains for energy optimization in \ac{TinyML} systems: (1) model-level compression offers immediate savings with accuracy trade-offs; (2) hardware--algorithm co-design unlocks efficiencies beyond software; (3) protocol- and system-level strategies manage communication, resource allocation, {duty cycling,} and local--cloud execution; and (4) memory- and data-centric optimizations reduce costly data movement. These findings provide both a research roadmap and practical guidance for developing energy-efficient \ac{TinyML} solutions in constrained environments.

Although the taxonomy captures the main optimization strategies reported in the selected literature, some studies did not propose or apply an explicit energy optimization technique. Instead, they focused on evaluating \ac{TinyML}-based solutions, anomaly detection pipelines, privacy-preserving \ac{IoT} frameworks, or edge-aware monitoring systems in which energy consumption was considered as part of the broader analysis \cite{al2023tiny,sanli2025privacy,alshuhail2025machine}. Therefore, these works were not assigned to one of the four optimization categories, but they remain relevant to the overall mapping because they contribute evidence on energy-aware \ac{TinyML} deployment scenarios.

% This taxonomy reveals four complementary domains for energy optimization in \ac{TinyML} systems: (1) model-level compression offers immediate savings with accuracy trade-offs; (2) hardware–algorithm co-design unlocks efficiencies beyond software; (3) protocol-level innovations are essential to manage communication overhead; and (4) memory-centric optimizations reduce costly data movement. 
% %The additional group of studies not directly mapped to the taxonomy also shows that part of the literature still addresses energy consumption mainly as an evaluation concern rather than as the target of an explicit optimization strategy. 
% These findings provide both a research roadmap and practical guidance for developing energy-efficient \ac{TinyML} solutions in constrained environments.

\subsection{Practical Insights and Implications}

In addition to addressing the research questions, the results 
of this systematic mapping study yield several actionable 
insights for practitioners and researchers engaged in the 
design and deployment of energy-efficient \ac{TinyML} solutions. 

First, the relative frequency and demonstrated impact of 
model-level optimizations such as post-training quantization, 
structured pruning, and knowledge distillation, provide 
guidance on which techniques to prioritize when balancing 
inference accuracy against energy budgets. 
For instance, 
{quantization reduced power consumption by 25--30\% on the same Cortex-M4 nodes with an average accuracy loss of about 1.5\%~\cite{villegas2024optimizing}}, 
showing clear applicability in battery-constrained 
devices.
Similarly, pruning combined with quantization reduced 
model size by up to 98.8\%~\cite{mnif2024combinative}, which 
is crucial for {edge devices with severe memory constraints, such as wearable and gesture-recognition systems}.

Second, communication optimizations demonstrated tangible 
impact. {As discussed in Section~4.5.2, in several studies communication accounted for a larger share of the energy budget than the inference itself~\cite{cortesi2024proximity,raza2023glass,villegas2024optimizing}. In practice, this means that the energy profile of the whole system should be measured before optimizing the model, since} transmission protocols and scheduling strategies can 
significantly extend device lifetime~\cite{sabovic2023towards}, 
sometimes more than model-level optimization alone. Third, 
the findings emphasize the role of hardware–algorithm 
co-design. Architectures such as GAP8 and \ac{CLADIMC} 
consistently achieved lower energy per inference than 
conventional \acp{MCU}~\cite{bian2022exploring,giordano2024tinyforge} 
by leveraging {parallel execution and optimized memory hierarchies}. This type of co-design is directly 
relevant for industrial automation and smart city 
applications, where system autonomy and reliability 
are essential.

Finally, by cataloging the diversity of metrics and reporting 
conventions found in the literature, this work highlights the 
need for standardized experimental 
protocols~\cite{bartoli2025benchmarking}. Establishing common 
definitions for energy per inference, end-to-end latency, and 
memory footprint will facilitate direct comparisons of future 
proposals. Moreover, the identification of under-explored 
areas such as automated model selection via \ac{AutoML}, 
memory-level caching strategies, and intelligently scheduled 
data transmission, points to promising directions for tool 
and framework development that can further accelerate the 
adoption of \ac{TinyML} in energy-sensitive applications.

\section{Open Challenges}
\label{open}

\ac{TinyML} enables machine learning inference on resource-constrained devices, but energy-aware deployment remains limited by methodological, algorithmic, and hardware-related challenges. Based on the analysis conducted in \acp{RQ} 1--4, the following research gaps and future research directions were identified. %Table~\ref{tab:open_challenges} summarizes these challenges by highlighting the current limitations and corresponding future research directions.

\begin{enumerate}
    \item \textbf{Lack of standardized energy measurement and benchmarking protocols} -- 
    The absence of standardized datasets, workloads, measurement setups, and reporting protocols hinders reproducibility and prevents objective comparison across studies. Although benchmark initiatives such as MLPerf Tiny~\cite{banbury2021mlperf} provide reference workloads for embedded inference, energy-aware \ac{TinyML} evaluation still lacks consistent procedures for reporting measurement conditions, hardware configurations, inference phases, and energy metrics. Future work should define benchmark suites and reporting guidelines that explicitly capture energy consumption, latency, memory usage, and accuracy under comparable conditions.

    \item \textbf{Energy--accuracy--latency trade-offs} -- 
    \ac{TinyML} deployments require balancing predictive performance, inference latency, memory footprint, and energy consumption. More complex models may improve accuracy but often increase computational cost and energy demand, while aggressive compression can reduce energy consumption at the expense of accuracy. Future research should develop multi-objective optimization methods that jointly evaluate accuracy, energy per inference, latency, and memory usage across different devices and application domains.

    \item \textbf{Energy impact of model compression and combined optimization pipelines} -- 
    Quantization, pruning, knowledge distillation, and architectural simplification are widely used to reduce model size and computational requirements. However, these techniques are often evaluated individually, and their combined effects on energy consumption, accuracy, latency, and hardware behavior are still not fully understood. Future studies should investigate integrated optimization pipelines and report how each optimization step affects energy consumption on real embedded platforms.

    \item \textbf{Hardware-dependent behavior and portability across architectures} -- 
    Energy consumption in \ac{TinyML} is strongly influenced by the target hardware, including \ac{MCU} architecture, memory hierarchy, clock frequency, accelerators, sensors, and communication interfaces. A model that is energy-efficient on one platform may not present the same behavior on another. Portable \ac{TinyML} solutions that generalize across heterogeneous hardware without extensive manual tuning remain limited. Future work should explore hardware-aware and hardware-adaptive deployment methods.

    \item \textbf{Automation in model selection and energy-aware tuning} -- 
    Most \ac{TinyML} models still require manual design, compression, and tuning to satisfy device constraints. Automated model selection, \ac{TinyML}-oriented \ac{AutoML}, and hardware-aware neural architecture search can reduce this effort, but these methods need to explicitly include energy, memory, and latency constraints rather than optimizing mainly for accuracy or model size.

    \item \textbf{Communication-aware \ac{TinyML} deployment} -- 
    Although local inference reduces dependence on cloud processing, many \ac{TinyML} systems still require communication for updates, event reporting, federated learning, or coordination within \ac{IoT} networks. Communication may dominate energy consumption depending on data rate, protocol, distance, and duty cycle. Future research should investigate adaptive communication strategies that jointly optimize sensing, inference, transmission, and sleep scheduling.

    \item \textbf{Reliability under real-world operating conditions} -- 
    \ac{TinyML} systems are often deployed in environments with unstable power supply, intermittent connectivity, sensor noise, temperature variation, hardware aging, and changing input distributions. These conditions can affect both inference quality and energy consumption. More research is needed on robust \ac{TinyML} models and deployment strategies that maintain reliable behavior under realistic operating constraints.

    \item \textbf{Learning with limited data and constrained update mechanisms} -- 
    Many \ac{TinyML} applications operate in domains where labeled data are scarce, user-specific patterns vary over time, or retraining is difficult after deployment. Federated learning, incremental learning, transfer learning, and synthetic data generation are possible alternatives, but their energy cost and feasibility on constrained devices remain insufficiently characterized. Future work should evaluate not only model accuracy but also the energy and memory overhead introduced by these learning strategies.

    \item \textbf{Underexplored low-power model families} -- 
    Most selected studies rely on \acp{CNN}, lightweight neural networks, or classical machine learning models. Emerging approaches such as Spiking Neural Networks (\ac{SNN}), event-driven models, neuromorphic computing, and intermittent \ac{TinyML} remain underexplored in energy-focused \ac{TinyML} evaluations. These approaches may offer advantages for ultra-low-power inference, but their practical benefits need to be validated using comparable benchmarks and real hardware measurements.
\end{enumerate}

Table~\ref{tab:open_challenges} provides a compact summary of the challenges discussed above, linking each challenge area to the main limitations observed in the literature and to possible future research directions.

\begin{table}[!]
\centering
\caption{Open challenges and future research directions for energy-aware \ac{TinyML}.}
\label{tab:open_challenges}
\scriptsize
\setlength{\tabcolsep}{2.5pt}
\renewcommand{\arraystretch}{1.05}
\begin{tabularx}{\textwidth}{p{3.1cm} X X}
\toprule
\textbf{Challenge area} & \textbf{Current limitations} & \textbf{Future research directions} \\
\midrule

Standardized energy measurement and benchmarking
& Lack of common datasets, workloads, measurement setups, hardware configurations, and reporting protocols, which hinders reproducibility and comparison across studies.
& Define benchmark suites and reporting guidelines that jointly capture energy consumption, latency, memory usage, accuracy, hardware configuration, and measurement conditions. \\

Energy--accuracy--latency trade-offs
& More complex models may improve accuracy but usually increase computational cost, latency, memory footprint, and energy consumption.
& Develop multi-objective optimization methods that explicitly balance energy per inference, accuracy, latency, and memory usage across devices and applications. \\

Model compression and integrated optimization pipelines
& Quantization, pruning, knowledge distillation, and architectural simplification are often evaluated separately, making their combined energy impact unclear.
& Investigate integrated optimization pipelines and quantify how each optimization step affects energy consumption, accuracy, latency, and hardware behavior. \\

Hardware-dependent behavior and portability
& Energy consumption varies across \ac{MCU} architectures, accelerators, memory hierarchies, clock frequencies, sensors, and communication interfaces.
& Develop hardware-aware and hardware-adaptive deployment methods that improve portability across heterogeneous platforms and reduce manual tuning. \\

Automation in model selection and energy-aware tuning
& \ac{TinyML} models often require manual design, compression, and parameter tuning to satisfy device-specific constraints.
& Extend \ac{TinyML} \ac{AutoML} and hardware-aware neural architecture search methods to include energy, memory, and latency constraints during model selection. \\

Communication-aware \ac{TinyML} deployment
& Communication for event reporting, cloud interaction, model updates, federated learning, or \ac{IoT} coordination may dominate energy consumption.
& Jointly optimize sensing, inference, transmission, duty cycling, and sleep scheduling in \ac{TinyML}-\ac{IoT} deployments. \\

Reliability under real-world operating conditions
& Sensor noise, unstable power supply, intermittent connectivity, temperature variation, hardware aging, and distribution shifts may affect both inference quality and energy behavior.
& Design robust \ac{TinyML} models and deployment strategies validated under realistic operating conditions. \\

Learning with limited data and constrained updates
& Labeled data are often scarce, while retraining, personalization, or model updates after deployment may be costly for constrained devices.
& Evaluate transfer learning, federated learning, incremental learning, and synthetic data generation while accounting for their energy and memory overhead. \\

Underexplored low-power model families
& Most selected studies rely on \acp{CNN}, lightweight neural networks, or classical machine learning models.
& Investigate \acp{SNN}, event-driven models, neuromorphic computing, and intermittent \ac{TinyML} using comparable benchmarks and real hardware measurements. \\

\bottomrule
\end{tabularx}
\end{table}

\section{Limitations of the Mapping Study}
\label{lim} 

Systematic review studies commonly encounter limitations \cite{petersen2015guidelines, daSilva2012}. This research is not exempt from such limitations, which include restricted coverage due to the formulation of the research questions and the search process, potential publication and selection biases, and possible inaccuracies during data extraction. Nevertheless, efforts were made to minimize these impacts by following well-established guidelines to structure the research protocol. Below, we outline each limitation and the strategies adopted to mitigate them.

Regarding the definition of the research questions, it is challenging to cover all aspects related to energy consumption in \ac{TinyML} algorithms for resource-constrained devices. To reduce this risk, a preliminary literature review was conducted, along with discussions among the researchers to validate and refine the research questions. Although it is possible that some relevant aspects were not addressed, we aimed to include the main open issues in the field, and we believe the study has fulfilled the primary purpose of a Systematic Mapping: to categorize studies and identify trends.

With respect to the search and selection process, the field of energy consumption in \ac{TinyML} still lacks a standardized terminology. This may have resulted in the omission of articles that used alternative terms not covered by the defined search strategy. To mitigate this impact, several tests were conducted using different combinations of search strings until a result set was obtained that aligned more closely with the study’s objectives.

{A further limitation concerns the uneven behavior of the search string across digital libraries. Because search engines differ in syntax and in the fields they match, the same string yielded only two records in IEEE Xplore, where matching was restricted to metadata. This effect was mainly caused by the generic term ``measurement'', which is common in the body of papers but rare in their metadata. To mitigate it, we relied on Scopus, which indexes IEEE publications, and conducted a supplementary IEEE Xplore search without this term, screening its 19 records with the same criteria, which led to the inclusion of two additional primary studies. The verbatim strings, search fields, and screening decisions are available in the replication package. Even so, studies that describe energy measurement using different terminology may not have been retrieved.}

Selection and publication bias is also a significant limitation in Systematic Mapping studies, as works with positive results are more frequently published than those reporting challenges or negative outcomes, potentially skewing the sample analyzed \cite{Kitchenham2007}. Furthermore, the selection criteria applied can affect the representativeness of the included studies. To reduce these risks, more than one researcher participated in the article screening process, reviewed the references of the most relevant works, and established clear inclusion and exclusion criteria to ensure a broader and more accurate selection of studies focused on energy consumption in the context of \ac{TinyML}.

Lastly, regarding data extraction and classification, Silva et al.~\cite{daSilva2012} highlights that the interpretation of extracted data may vary among researchers, possibly resulting in inconsistent classifications. To address this issue, a cross-validation process was implemented, where at least two authors reviewed and categorized the studies, resolving any disagreements through consensus. This ensured greater rigor in the analysis of the results.

\section{Conclusions}
\label{con}

{This systematic mapping study investigated energy consumption in \ac{TinyML} algorithms deployed on resource-constrained devices, analyzing 25 primary studies published between 2020 and 2025. The analysis covered the algorithms evaluated, the methodologies and metrics used to assess energy consumption, the technical factors that influence it, and the optimization strategies reported in the literature.}

{Regarding the algorithms, convolutional architectures were the most frequent, appearing in 14 of the 25 studies, while lightweight neural networks and classical machine learning models were mainly adopted under stricter memory and energy constraints. Regarding evaluation, direct measurement with physical instruments was the most common methodology, identified in 12 studies, followed by analytical and inference-time-based estimations. However, the studies differed considerably in measurement setups and reported metrics, such as energy per inference, average power, and energy per work cycle, which limits direct comparison across studies.}

{The results also show that energy consumption is not determined by the algorithm alone, but by its interaction with the hardware platform, communication strategy, numerical precision, and timing parameters. The same model presented differences of more than one order of magnitude in energy consumption across platforms, and the fastest device was not necessarily the most energy-efficient. Consistently, the optimization strategies were organized into four complementary categories, namely model compression, hardware-aware design, system and communication optimization, and data and processing optimization, with hardware-aware and data-centric strategies achieving gains comparable to those of model compression. Nevertheless, few studies isolated the effect of each technique on the same device, which calls for caution when attributing energy savings to a specific optimization.}

{These findings point to several directions for future research. First, the field would benefit from standardized reporting of energy results, including energy per inference, measurement instruments, device configuration, and the execution phases considered. Second, optimization techniques should be evaluated individually and in combination on the same hardware, so that their energy effects can be separated from those of the platform. Third, automated model selection and neural architecture search should explicitly include energy constraints, and underexplored model families, such as spiking neural networks, should be evaluated with real hardware measurements. Finally, future updates of this mapping could adopt broader search strategies to capture studies that report energy measurements using different terminology.}

\section*{Declarations}

\noindent\textbf{Funding.}
This work was supported by the National Council for Scientific and
Technological Development (CNPq), Brazil, under grant number
401147/2025-8.

\medskip

\noindent\textbf{Conflict of Interest.}
The authors declare no conflicts of interest.

\medskip

\noindent\textbf{Author Contributions.}
{Clariele Almeida and Rafael Moura conducted the initial screening of titles and abstracts. Clariele Almeida, Renan Beserra, and Rafael Moura assessed the full texts against the eligibility criteria, extracted and analyzed the data, prepared the replication package, and wrote the original draft. They also contributed to the methodology and literature searches. Danilo Araújo and Sérgio Chevtchenko critically reviewed and improved the initial draft. Bruno Nogueira and Ermeson Andrade contributed to the interpretation of the findings and critically revised the manuscript to develop its final version. Ermeson Andrade provided methodological guidance and supervised all stages of the study. All authors read and approved the final manuscript.}

\medskip

\noindent\textbf{Data Availability.}
The data and materials supporting this study, including the study
protocol, search results, and extracted data, are openly available
in Zenodo at \url{https://zenodo.org/records/23348211}.

\bibliography{sn-bibliography}% common bib file
%% if required, the content of .bbl file can be included here once bbl is generated
%%\input sn-article.bbl

\end{document}